\documentclass{article}

\usepackage[T1]{fontenc}
\usepackage{microtype}
\usepackage{graphicx}
\usepackage{capt-of}
\usepackage{subfigure}
\usepackage{booktabs}
\usepackage[table]{xcolor}
\usepackage{xurl} 
\usepackage{hyperref}

\usepackage{amsmath}
\usepackage{amssymb}
\usepackage{mathtools}
\usepackage{xspace}
\usepackage{multirow}
\usepackage{array}
\usepackage{siunitx}
\usepackage{enumitem}
\usepackage[most]{tcolorbox}

\usepackage{tikz}
\usetikzlibrary{arrows.meta,positioning,fit,backgrounds,calc,shapes.geometric}

\newif\ifanonymoussubmission
\ifdefined\anonymoussubmission
  \usepackage{mlsys2025}
  \hypersetup{
    pdfauthor={Anonymous Authors},
    pdftitle={Understanding the Impact of Tolerance Threshold in Automated GPU Kernel Optimization},
  }
\else
  \anonymoussubmissionfalse
  \usepackage[accepted]{mlsys2025}
  \hypersetup{
    pdfauthor={Zirui Wang, Yunbo Ni, Jinjie Liu, Chijin Zhou, Shaohua Li},
    pdftitle={Understanding the Impact of Tolerance Threshold in Automated GPU Kernel Optimization},
  }
\fi

\AtBeginDocument{\mlsystitlerunning{Understanding Tolerance Threshold in Automated GPU Kernel Optimization}}

\providecommand{\Description}[2][]{}

\ifanonymoussubmission\else
  \newcommand{\mlsyscorrfootnote}[1]{%
    \let\mlsystitle@orig\mlsystitle
    \def\mlsystitle##1{\mlsystitle@orig{##1}%
      \global\let\mlsystitle\mlsystitle@orig
      \global\let\corrlegendtext##1}%
  }
  \fi

\newif\ifcameraready
\camerareadyfalse

\ifcameraready
  
\else
  
\fi

\ifcameraready
  \newcommand{\todo}[1]{}
\else
  \newcommand{\todo}[1]{\textcolor{red}{\textsf{\footnotesize[TODO: #1]}}}
\fi

\ifcameraready
  
\else
  
\fi

\newcommand{\placeholderfig}[2]{%
  \begin{tikzpicture}
    \node[draw, dashed, line width=0.6pt, color=black!45,
          minimum height=#1\baselineskip,
          text width=\dimexpr\linewidth-2.4em\relax,
          align=center, inner sep=0.8em]
      {\textsf{\footnotesize\textcolor{black!60}{PLACEHOLDER\\[0.3em]#2}}};
  \end{tikzpicture}%
}

\newcommand{\fpthirtytwo}{\mbox{FP32}\xspace}
\newcommand{\fpsixteen}{\mbox{FP16}\xspace}
\newcommand{\bfsixteen}{\mbox{BF16}\xspace}

\newcommand{\figref}[1]{Figure~\ref{#1}}
\newcommand{\tabref}[1]{Table~\ref{#1}}
\newcommand{\secref}[1]{\S\ref{#1}}

\newcommand{\findingbox}[2]{%
\begin{tcolorbox}[
  enhanced,
  width=\linewidth,
  colframe=black!65,
  arc=1pt,
  left=2pt,
  right=2pt,
  top=1pt,
  bottom=1pt,
  boxsep=1pt,
  before skip=6pt,
  after skip=6pt
]
\textbf{#1.} #2
\end{tcolorbox}%
}

\usepackage{xspace}
\usepackage{pifont}
\usepackage{amsmath}

\newcommand{\para}[1]{\noindent\textbf{#1}\xspace}

\newcommand{\cuhk}{The Chinese University of Hong Kong}
\newcommand{\ecnu}{East China Normal University}
\newcommand{\affilcuhk}{cuhk}
\newcommand{\affilecnu}{ecnu}

\begin{document}
\twocolumn[
    \mlsystitle{Understanding the Impact of Tolerance Threshold \\in Automated GPU Kernel Optimization}
    \begin{mlsysauthorlist}
      \mlsysauthor{Zirui Wang}{\affilcuhk}
      \mlsysauthor{Yunbo Ni}{\affilcuhk}
      \mlsysauthor{Jinjie Liu}{\affilcuhk}
      \mlsysauthor{Chijin Zhou}{\affilecnu}
      \mlsysauthor{Shaohua Li}{\affilcuhk}
    \end{mlsysauthorlist}

    \mlsysaffiliation{\affilcuhk}{\cuhk}
    \mlsysaffiliation{\affilecnu}{\ecnu}

    \mlsyscorrespondingauthor{Shaohua Li}{shaohuali@cuhk.edu.hk}

    \vskip 0.3in
    \begin{abstract}
GPU kernel optimization is increasingly driven by standardized benchmarks and autonomous kernel agents. 
To assess correctness, these benchmarks and agents commonly use differential testing, accepting an optimized kernel when its outputs remain within a predefined tolerance threshold of the reference
outputs.
However, how discrepancies accepted by these tests affect model behavior during inference remains poorly understood.
In this paper, we investigate this question through real kernel replacements and controlled perturbations.
We first integrate \textbf{1,160} published optimized kernel implementations from CUDA-L1 and the KernelBench leaderboard into 13 real-world models across three numerical precisions.
Across 2,739 replacement evaluations, \textbf{55.28\%} produce at least one output change. Although accuracy changes are usually small, 11 evaluations incur accuracy drops greater than 50\%. In the most severe case, Gemma-4-E4B's accuracy on MMLU \textbf{falls from 67.0\% to 5.5\%}.
We then conduct 6,928 controlled perturbation evaluations to examine how numerical discrepancies affect model behavior.
The effects depend on numerical precision, operator, and dataset. For Gemma-4-12B on MMLU in \fpsixteen{}, perturbations of the same magnitude change outputs for \textbf{73.6\%} of model inputs when applied to \texttt{post\_attention\_rmsnorm}, but cause \textbf{no output changes} when applied to \texttt{lm\_head}.
Our findings show that passing kernel-level tests does not prevent substantial and varied changes in model behavior.
We hope our research findings can inspire the community to further explore correctness criteria in future kernel benchmarks and optimization agents.
\end{abstract}
]
\printAffiliationsAndNotice{}

    \section{Introduction}
\label{sec:intro}

GPU kernel performance directly affects the latency and throughput of model serving.
Autonomous kernel agents driven by large language models (LLMs) now generate and
iteratively optimize GPU kernels~\cite{dai2026cudaagent,dong2026kernelblaster,ye2026cake,li2026cudal1}.
Standardized benchmarks such as KernelBench evaluate these kernels on both runtime speedup and
correctness~\cite{ouyang2025kernelbench}.
A faster kernel is useful only if its effect on model behavior stays acceptable.

Current benchmarks check correctness by differential testing, which we call the
\emph{kernel-level test}: they run an optimized kernel and a reference
implementation on the same test inputs and compare the outputs.
Exact equality would reject valid optimizations that reorder floating-point
operations, such as parallel reductions.
Benchmarks therefore use absolute and relative \emph{tolerance thresholds}
to check whether the output tensors of the optimized and reference
kernels are close enough.
We call the elementwise difference between these tensors the
\emph{kernel-level discrepancy}.
KernelBench proposed heuristic per-precision tolerance thresholds, $10^{-4}$ for \fpthirtytwo and $10^{-2}$ for \fpsixteen and \bfsixteen, which have become a common standard in the community~\cite{ouyang2025kernelbench}. These values were chosen empirically, with limited evidence of their implications for model inference.

\begin{figure}[t]
  \centering
  \includegraphics[width=\columnwidth]{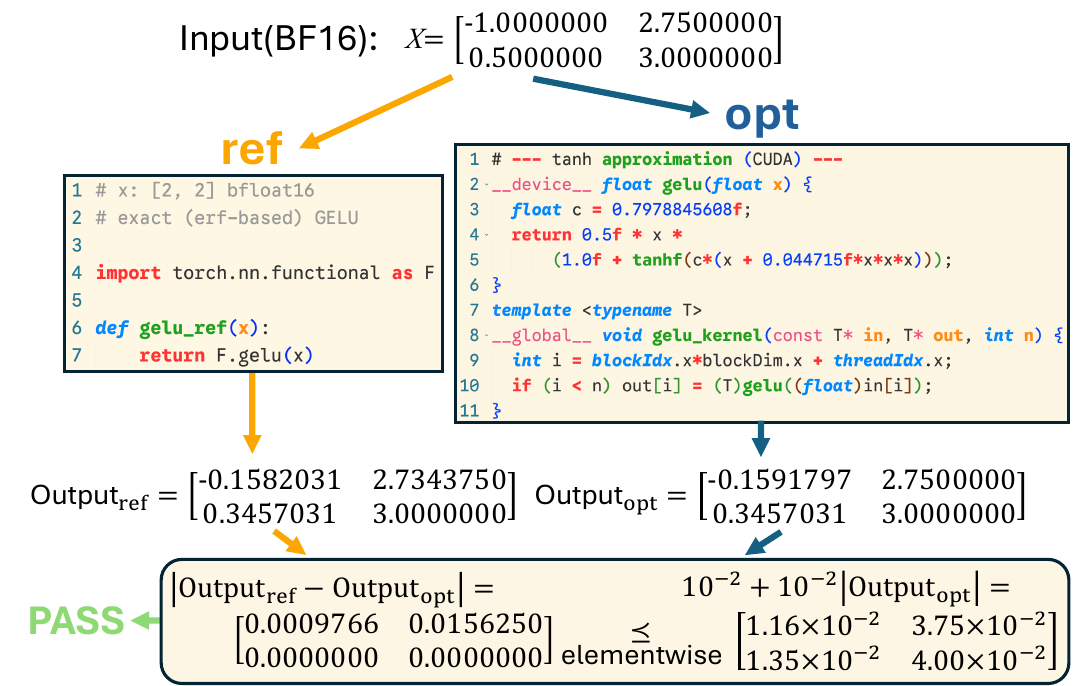}
  \vspace{-20pt}
  \caption{An optimized GELU kernel passes the kernel-level test in \bfsixteen{}.}
  \label{fig:intro-tolerance-gap}
  \Description{Differential correctness check comparing a PyTorch reference GELU and an
  optimized CUDA GELU kernel on the same test input. Every element satisfies
  $\mathrm{atol}+\mathrm{rtol}|y|$, so the check passes.}
  \vspace{-10pt}
\end{figure}

\begin{figure}[t]
  \centering
  \IfFileExists{figures/assets/fig-motivating-example.pdf}{%
    \includegraphics[width=\columnwidth]{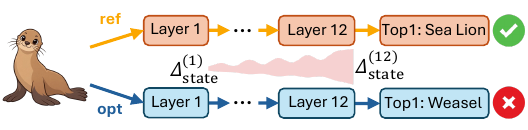}%
  }{%
    \placeholderfig{6}{%
      Top: the reference and optimized GELU receive the same operator input tensor and produce
      a small, nonzero kernel-level discrepancy that remains within the tolerance threshold.
      Bottom: the two execution paths pass through Layer 1, an
      ellipsis, and Layer 12. A non-monotonic band labeled ``propagated
      model-state discrepancy'' separates the paths before their top-1 outputs
      differ: weasel versus sea lion.}%
  }
  \vspace{-20pt}
  \caption{An optimized GELU kernel replacement changes ViT-Base's top-1 prediction.}
  \label{fig:motivating-example}
  \vspace{-10pt}
\end{figure}

\figref{fig:intro-tolerance-gap} illustrates this gap with a GELU kernel. The optimized kernel (right) accelerates the GELU computation by approximating the reference kernel (left). The kernel-level test accepts the optimized kernel because its output discrepancy is within the predefined tolerance threshold for \bfsixteen.
When this kernel replaces GELU throughout
ViT-Base~\cite{dosovitskiy2021vit}, however, the kernel-level discrepancy in
GELU outputs propagates through subsequent layers and alters
intermediate model states (\figref{fig:motivating-example}).
For the same input image, the top-1 prediction changes from
sea lion to weasel.
Passing the kernel-level test on probe inputs therefore does
not ensure unchanged model outputs.

This example raises a broader question: \emph{How do optimized kernels that pass kernel-level tests under the specified tolerance
thresholds affect model behavior during inference?}

We study 13 models, seven task types, and 49 operators
(\secref{sec:methodology}) through two complementary evaluations.
\begin{itemize}[wide, itemsep=1pt, topsep=1pt, parsep=0pt, partopsep=0pt]
  \item \textbf{Real kernel replacements (\secref{sec:replacement}).}
  We integrate 1,160 optimized kernel implementations into 13 real-world models across three numerical precisions, using published kernels from CUDA-L1~\cite{li2026cudal1} and the KernelBench leaderboard~\cite{ouyang2025kernelbench} that pass the kernel-level test.
  Across 2,739 replacement evaluations, 55.28\% produce at least
  one output change and 41.66\% produce at least one negative flip, where a correct reference output becomes incorrect after replacement.
  Output changes are common across evaluations but usually  sparse across model inputs.
  Accuracy changes remain within $\pm 0.5\%$ in 94.7\% of evaluations and only 11 incur accuracy drops greater than 50\%, while the largest observed drop is 61.5\%, reducing Gemma-4-E4B's accuracy on MMLU from 67.0\% to 5.5\%. Even under the same inference configuration, optimized kernels that pass the kernel-level test can have sharply different effects:
  for Gemma-4-E4B on MMLU in \fpsixteen{}, one \texttt{lm\_head} kernel leaves accuracy unchanged, while two others each cause a 59.5\% accuracy drop. We further analyze variation across configurations and kernels and identify the most affected model inputs.

  \item \textbf{Controlled kernel perturbations (\secref{sec:perturbation}).}
  We inject perturbations into the output tensors of reference kernels and vary their magnitude across 6,928 evaluations.
  Larger perturbations generally cause more output changes and negative flips.
  However, in 670 of 1,299 configurations, increasing the perturbation magnitude sometimes produces fewer output changes.
  The effects also depend on numerical precision, operator, and dataset.
  For Gemma-4-12B on MMLU in \fpsixteen{}, perturbations of the same magnitude change outputs for 73.6\% of model
  inputs when applied to \texttt{post\_attention\_rmsnorm},
  but cause no output changes when applied to \texttt{lm\_head}.
  We further examine how these effects vary across configurations and how perturbation results relate to real replacement effects.
\end{itemize}

    \section{Evaluation Design}
\label{sec:methodology}

\para{Study overview.}
We study how kernel-level discrepancies affect model behavior through real kernel replacements and controlled kernel perturbations.
An \emph{inference configuration} consists of a model, a selected operator $p$, a numerical precision, and an evaluation dataset.
In the replacement study, a \emph{replacement evaluation} compares reference inference with inference using one optimized kernel that passes the kernel-level test, under the same inference configuration (\secref{sec:replacement}).
In the perturbation study, a \emph{perturbation evaluation} compares reference inference with inference that injects perturbations of a specified magnitude into the output tensors of the reference kernel, under the same inference configuration (\secref{sec:perturbation}).
In each evaluation, we compare the two inferences on the same model inputs and measure output changes, negative flips, and accuracy change.

\para{Operators and kernels.}
An \emph{operator} specifies a computation in a model, and a \emph{kernel} implements that computation on a device.
In \figref{fig:replacement-example}, ReLU is the operator. Panels (a) and (b) show its reference implementation in PyTorch and optimized implementation in CUDA.
We use $p$ to denote the selected operator, and $K_p^{\mathrm{ref}}$ and $K_p^{\mathrm{opt}}$ to denote its reference and optimized kernels. An operator may be invoked multiple times during one forward pass.

\subsection{Models, Kernels, and Datasets}
\label{sec:methodology:coverage}

We evaluate 13 open-source models across seven task types:
multiple-choice QA, text generation, image classification, extractive
QA, text classification (NLI), ASR-CTC, and ASR-seq2seq
(\tabref{tab:models-workloads})~\cite{yang2025qwen3,gemmateam2026gemma4,dosovitskiy2021vit,oquab2024dinov2,liu2022convnext,radford2021clip,he2023debertav3,baevski2020wav2vec2,radford2023whisper,hendrycks2021mmlu,cobbe2021gsm8k,recht2019imagenetv2,hendrycks2021imageneta,wang2019imagenetsketch,rajpurkar2018squad2,bartolo2020adversarialqa,williams2018mnli,bowman2015snli,nie2020anli,panayotov2015librispeech}.
For each dataset we sample a subset once and reuse it for
every reference, replacement, and perturbation evaluation on it.
Each evaluation runs inference on every model input in that subset and produces
one final output per model input.

\begin{table}[t]
  \centering
  \caption{Evaluated task types, models, and datasets.}
  \label{tab:models-workloads}
  \scriptsize
  \setlength{\tabcolsep}{2.5pt}
  \renewcommand{\arraystretch}{1.08}
  \begin{tabular}{@{}>{\raggedright\arraybackslash}p{1.8cm}>{\raggedright\arraybackslash}p{3.1cm}>{\raggedright\arraybackslash}p{1.45cm}r@{}}
    \toprule
    \textbf{Task type} & \textbf{Models (ID)} & \textbf{Datasets} & \textbf{\#Samples} \\
    \midrule
    Multiple-choice\newline QA
      & Qwen3-8B/14B (Q8, Q14),\newline
        Gemma-4-E4B/12B (G4, G12)
      & MMLU
      & 1,000 \\
    \midrule
    Text generation
      & Qwen3-8B/14B (Q8, Q14),\newline
        Gemma-4-E4B/12B (G4, G12)
      & GSM8K
      & 1,000 \\
    \midrule
    Image classification
      & ViT-Base/Large (VB, VL),\newline
        DINOv2-Large (DINO),\newline
        ConvNeXT-Base (CN),\newline
        CLIP ViT-B/32 (CLIP)
      & ImageNet-V2,\newline
        ImageNet-A,\newline
        ImageNet-\newline Sketch
      & 3,000 \\
    \midrule
    Extractive\newline QA
      & DeBERTa-v3-L (QA)
      & SQuAD~v2,\newline
        AdvQA
      & \begin{tabular}[t]{@{}r@{}} 3,612 \\ 3,000 \end{tabular} \\
    \midrule
    Text classification \newline (NLI)
      & DeBERTa-v3-L (NLI)
      & MNLI, SNLI, \newline
      ANLI
      & 2,000 \\
    \midrule
    ASR-CTC
      & Wav2Vec2-base (W2V)
      & LibriSpeech\newline (clean/other)
      & \begin{tabular}[t]{@{}r@{}}256\\256\end{tabular} \\
    \midrule
    ASR-seq2seq
      & Whisper-small (WSP)
      & LibriSpeech\newline (clean/other)
      & \begin{tabular}[t]{@{}r@{}}256\\256\end{tabular} \\
    \bottomrule
  \end{tabular}
  \vspace{-10pt}
\end{table}

We take optimized kernels from the official CUDA-L1 artifacts and the
public KernelBench leaderboard~\cite{li2026cudal1,ouyang2025kernelbench}.
A kernel enters the replacement study only if it passes the kernel-level
test at the target precision (\secref{sec:methodology:setup}) and completes
model inference without fallback or runtime failures.
All studies cover the same 49 operators, as listed in \tabref{tab:replacement-matrix}.
For each operator in each model, the table lists the number of distinct
optimized kernels included in the replacement study, in the order
[\fpthirtytwo{}/\fpsixteen{}/\bfsixteen{}].
A zero means that no optimized kernel meets the inclusion criteria
for that model and operator at that precision.


\begin{table*}[t]
  \centering
  \caption{Operators and optimized kernel counts per model.}
  \label{tab:replacement-matrix}
  \scriptsize
  \setlength{\tabcolsep}{3pt}
  \ifdefined\rqonematrixrightcol\else\newlength{\rqonematrixrightcol}\fi
  \setlength{\rqonematrixrightcol}{8.7cm}
  \rowcolors{2}{white}{gray!15}
  \begin{tabular}{@{}>{\raggedright\arraybackslash}p{\dimexpr\textwidth-\rqonematrixrightcol-6pt\relax}>{\raggedright\arraybackslash}p{\rqonematrixrightcol}@{}}
    \toprule
    \textbf{Operators} & \textbf{Models [optimized kernel counts in \fpthirtytwo{}/\fpsixteen{}/\bfsixteen{}]} \\
    \midrule
    \texttt{attention\_\{bmm,softmax\}} & Q8, Q14, G4, G12, VB, VL, QA, NLI, WSP [3/3/3]; DINO, CLIP [1/1/1]; W2V [3/0/3] \\
    \texttt{attention\_out\_proj}, \texttt{\{key,query,value\}\_proj} & Q8, Q14, G4, G12, VB, VL [3/3/3] \\
    \texttt{attention\_\{pv,qk\}\_bmm} & Q8, Q14, G4, G12 [3/3/3]; VB, VL [4/4/4] \\
    \texttt{attention\_scale} & Q8, Q14, G4, G12, VB, VL, DINO, CLIP, QA, NLI, WSP [1/1/1]; W2V [1/0/1] \\
    \texttt{\{block,downsample,patch\}\_layernorm}, \texttt{conv2d}, \texttt{\{depthwise\_spatial,downsample\}\_conv2d}, \texttt{global\_avg\_pool}, \texttt{layer\_scale} & CN [1/1/1] \\
    \texttt{conv1d} & W2V [1/0/1] \\
    \texttt{encoder\_conv1d} & WSP [1/1/1] \\
    \texttt{feature\_l2norm}, \texttt{logit\_scale}, \texttt{quick\_gelu} & CLIP [1/1/1] \\
    \texttt{final\_layernorm} & VB, VL, CN [1/1/1] \\
    \texttt{\{final,input,k,post\_attention,q\}\_rmsnorm}, \texttt{rmsnorm} & Q8, Q14, G4, G12 [1/1/1] \\
    \texttt{gelu} & VB, VL, WSP [0/3/3]; DINO [1/1/1]; W2V [0/0/3] \\
    \texttt{hidden\_linear} & CLIP [3/3/0] \\
    \texttt{layernorm} & VB, VL, DINO, CN, CLIP, QA, NLI [1/1/1]; W2V [3/0/3]; WSP [3/3/3] \\
    \texttt{linear} & Q8, Q14, G4, G12, VB, VL, CN, QA, NLI, WSP [3/3/3]; DINO [3/3/0]; W2V [3/0/3] \\
    \texttt{lm\_head}, \texttt{mlp\_\{down,gate,up\}\_proj} & Q8, Q14, G4, G12 [3/3/3] \\
    \texttt{mlp\_activation} & Q8, Q14, G4, G12 [3/3/3]; CN [1/1/1]; QA, NLI [1/2/2] \\
    \texttt{mlp\_fc\{1,2\}} & VB, VL [3/3/3] \\
    \texttt{norm\{1,2\}\_layernorm} & VB, VL [1/1/1] \\
    \texttt{patch\_conv} & VB, VL [2/2/2]; CLIP [1/1/1] \\
    \texttt{patch\_conv2d} & DINO, CN [1/1/1] \\
    \texttt{\{post,pre\}\_feedforward\_rmsnorm}, \texttt{v\_rmsnorm} & G4, G12 [1/1/1] \\
    \texttt{relative\_attention\_bmm}, \texttt{task\_head} & QA, NLI [3/3/3] \\
    \bottomrule
  \end{tabular}
  \vspace{-16pt}
\end{table*}

Our study covers 1,299 inference configurations, comprising
2,739 replacement evaluations and 6,928 perturbation evaluations.
The counts differ because a configuration can include multiple optimized
kernels in the replacement study and multiple perturbation magnitudes in
the perturbation study.

\subsection{Measurements \& Metrics}
\label{sec:methodology:metrics}
We define three evaluation metrics to measure the impact of an optimized kernel.

\para{Kernel-level discrepancy.}
The differential test compares the reference and optimized kernels on the
same test inputs. For one such test input, let $y_{\mathrm{ref}}$ and
$y_{\mathrm{opt}}$ be the output tensors of the reference and
optimized kernels.
Their kernel-level discrepancy is the element-wise difference
$\left|y_{\mathrm{opt}}-y_{\mathrm{ref}}\right|$.
The kernel-level test uses \texttt{torch.allclose} with the
reference output first and the optimized output second, and requires
every element $j$ to satisfy
\[
\left|y_{\mathrm{opt},j}-y_{\mathrm{ref},j}\right|
\leq \mathrm{atol}+\mathrm{rtol}\left|y_{\mathrm{opt},j}\right|.
\]
Here $\mathrm{atol}$ and $\mathrm{rtol}$ are the predefined absolute and
relative tolerance thresholds.

\para{Output change \& negative flip.}
An output change occurs when inference after a kernel replacement or
an injected perturbation produces a different final output from
the reference inference on the same model input.
Different tasks require different ways of evaluating output changes.
We compare the selected option on MMLU, the parsed numerical answer on
GSM8K, the top-1 label on ImageNet-V2/A/Sketch, the extracted span on
SQuAD~v2 and AdvQA, the predicted label on NLI, and the decoded
transcript on LibriSpeech.
A negative flip occurs when the reference output is correct and the
output after replacement or perturbation is not.
For every model input, correctness requires an exact match to the ground truth, not token-level F1 or word error rate.

For an evaluation of $N$ model inputs, let $n_{\mathrm{change}}$ and
$n_{\mathrm{flip}}$ be the numbers of output changes and negative
flips.
The rates are
\[
R_{\mathrm{change}}=\frac{n_{\mathrm{change}}}{N},\qquad
R_{\mathrm{flip}}=\frac{n_{\mathrm{flip}}}{N}.
\]

\para{Accuracy change.}
Let $\mathrm{Acc}^{\mathrm{ref}}$ be the accuracy of reference inference,
computed as the number of model inputs with correct final outputs divided
by the total number of evaluated model inputs, and let $\mathrm{Acc}$ be the
corresponding fraction after replacement or perturbation.
We define
$\Delta_{\mathrm{acc}}=\mathrm{Acc}-\mathrm{Acc}^{\mathrm{ref}}$.
A positive $\Delta_{\mathrm{acc}}$ is an accuracy gain and a negative value means an accuracy drop. We report accuracy differences in percent.

\subsection{Experimental Setup}
\label{sec:methodology:setup}

\para{Hardware \& software.}
All experiments run in a Linux container on one NVIDIA RTX PRO 6000
Blackwell Server Edition GPU, with NVIDIA driver 580.95.05, CUDA
Toolkit 13.0, and PyTorch 2.11.0 built with CUDA 12.8.

\para{Precision \& tolerance thresholds.}
We evaluate \fpthirtytwo, \fpsixteen, and \bfsixteen.
The kernel-level test uses $\mathrm{atol}=\mathrm{rtol}=10^{-4}$
for \fpthirtytwo and $\mathrm{atol}=\mathrm{rtol}=10^{-2}$ for
\fpsixteen and \bfsixteen.
A replacement kernel must pass this test at the precision used in
inference.

\para{Decoding.}
Within each evaluation, the two inferences share model parameters,
precision, preprocessing, and decoding.
Language-model generation (GSM8K) and Whisper decoding are greedy
(\texttt{do\_sample=False}, \texttt{num\_beams=1}).
For MMLU, we select among A/B/C/D using next-token scores.
Wav2Vec2 uses CTC argmax decoding. Other tasks use the same task-specific decoding for the reference
inference and for inference after replacement or perturbation.
    \section{Real Kernel Replacement}
\label{sec:replacement}

We study how optimized kernels that pass the kernel-level test
affect model behavior under the evaluation design in
\secref{sec:methodology}.
We address the following research questions:

\begin{itemize}[labelwidth=!, labelindent=5pt, itemsep=1pt, topsep=1pt, parsep=0pt, partopsep=0pt, leftmargin=15pt]
    \item \textbf{RQ3.1 (Effects on model behavior).}
    How often do these replacements produce an output change or a
    negative flip, what fraction of final outputs do they affect, and how much does accuracy change?
 
    \item \textbf{RQ3.2 (Variation across configurations and optimized kernels).}
    How do replacement effects vary across inference configurations and
    optimized kernels, and why can different optimized kernels that pass the
    kernel-level test still produce substantially different model behavior
    under the same inference configuration?

    \item \textbf{RQ3.3 (Characteristics of affected model inputs).}
    Are output changes concentrated on low-margin model inputs, and does
    this pattern hold for both small and large changes in model behavior?

\end{itemize}

\para{Replacement procedure.}
\label{sec:replacement:procedure}
Let $K^{\mathrm{ref}}_p$ and $K^{\mathrm{opt}}_p$ be the reference and
optimized kernels for operator $p$.
At every invocation in one forward pass, $p$ uses $K^{\mathrm{opt}}_p$ instead of $K^{\mathrm{ref}}_p$.
For an invocation on model input $x$, let $F_{<p}(x)$
produce $p$'s input tensor and $F_{>p}$ consume its output:
\[
\begin{aligned}
F^{\mathrm{ref}}(x)&=F_{>p}\bigl(K^{\mathrm{ref}}_p(F_{<p}(x))\bigr), \\
F^{\mathrm{opt}}_p(x)&=F_{>p}\bigl(K^{\mathrm{opt}}_p(F_{<p}(x))\bigr).
\end{aligned}
\]
The two inferences share model parameters and differ only in the
kernel for $p$. 
All models run in eager mode without \texttt{torch.compile}, so operator
boundaries remain visible.
For a standalone module such as Linear, RMSNorm, or GELU, we replace
the module with a wrapper that keeps the original parameters and uses
$K^{\mathrm{opt}}_p$ on every invocation of $p$.
For fused operators such as Attention, suboperations remain fused even
in eager mode. We register a custom attention function to replace only
the selected suboperation, such as Softmax or the QK matmul.

The wrapper records each invocation of $p$ and whether
$K^{\mathrm{opt}}_p$ or $K^{\mathrm{ref}}_p$ ran.
We keep a replacement evaluation only if every recorded invocation uses
$K^{\mathrm{opt}}_p$ without fallback.
Which kernels enter the study is defined in
\secref{sec:methodology}.

\begin{figure*}[t]
  \centering
  \begin{minipage}[t]{0.64\textwidth}
    \vspace{0pt}
    \centering
    \includegraphics[width=\linewidth]{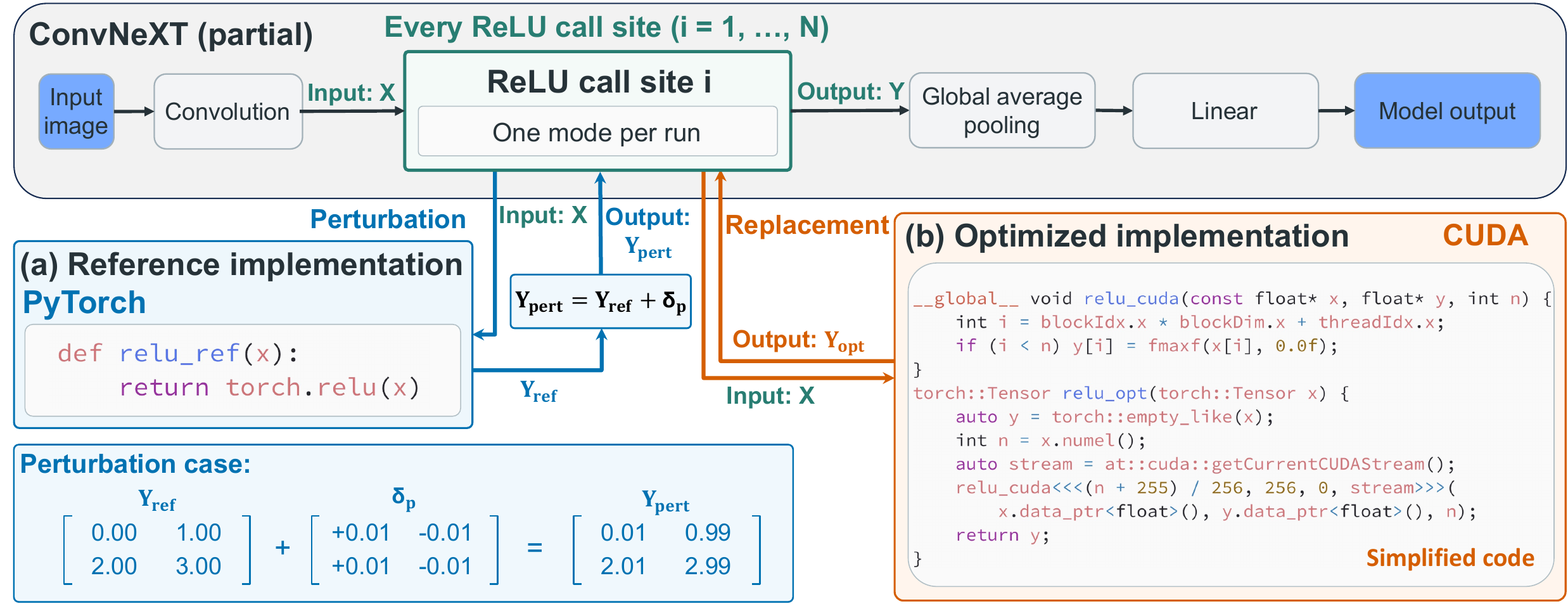}
    \vspace{-20pt}
    \captionof{figure}{Example of kernel replacement and controlled perturbation for a ReLU operator in ConvNeXT.}
    \label{fig:replacement-example}
  \end{minipage}\hfill
  \begin{minipage}[t]{0.34\textwidth}
    \vspace{0pt}
    \centering
    \includegraphics[width=\linewidth]{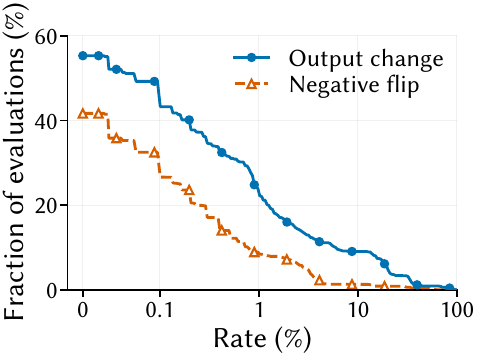}
    \vspace{-24pt}
    \captionof{figure}{Fraction of evaluations with $R_{\mathrm{change}}$ or
      $R_{\mathrm{flip}}$ above the x-axis value.}
    \label{fig:replacement-overall-a}
  \end{minipage}
  \vspace{-14pt}
\end{figure*}

\para{Example.}
In \figref{fig:replacement-example}, a ConvNeXT ReLU call lies between
convolution and global average pooling. The wrapper replaces the PyTorch
reference kernel in panel (a) with the optimized CUDA kernel in panel (b),
passing input tensor $X$ to it and returning $Y_{\mathrm{opt}}$ to the
remaining computation. All other operations, parameters, and model inputs
remain unchanged.

\subsection{Effects on Model Behavior}
\label{sec:replacement:rq31}

\begin{table*}[!htp]
  \centering
  \caption{Replacement evaluations across models.}
  \label{tab:replacement-summary}
  \scriptsize
  \setlength{\tabcolsep}{2pt}
  \begin{tabular*}{\textwidth}{@{\extracolsep{\fill}}llrrrrrrrrrrrrrr@{}}
    \toprule
    \textbf{Effect} & \textbf{Statistic} & \textbf{Q8} & \textbf{Q14} &
    \textbf{G4} & \textbf{G12} & \textbf{VB} & \textbf{VL} & \textbf{DINO} &
    \textbf{CN} & \textbf{CLIP} & \textbf{QA} & \textbf{NLI} & \textbf{W2V} &
    \textbf{WSP} & \textbf{Overall} \\
    \midrule
    & \# Evaluations ($N$)
      & 294 & 294 & 312 & 312
      & 396 & 396 & 72 & 135
      & 90 & 112 & 168 & 62
      & 96 & \textbf{2,739} \\
    \midrule
    \multirow{3}{*}{Output change}
      & Affected evaluations (\%)
      & 60.88 & 58.50 & 65.38 & 64.42
      & 59.60 & 62.63 & 38.89 & 25.93
      & 36.67 & 50.89 & 24.40 & 56.45
      & 46.88 & \textbf{55.28} \\
      & Median $R_{\mathrm{change}}$ (\%)
      & 0.20 & 0.20 & 0.40 & 0.30
      & 0.03 & 0.10 & 0 & 0
      & 0 & 0.03 & 0 & 10.55
      & 0 & \textbf{0.07} \\
      & P90 $R_{\mathrm{change}}$ (\%)
      & 17.80 & 13.80 & 33.39 & 19.60
      & 0.87 & 1.08 & 0.27 & 68.67
      & 1.06 & 0.77 & 0.10 & 42.85
      & 4.10 & \textbf{5.92} \\
    \midrule
    \multirow{3}{*}{Negative flip}
      & Affected evaluations (\%)
      & 51.36 & 50.68 & 58.97 & 57.05
      & 34.60 & 45.96 & 34.72 & 20.74
      & 16.67 & 31.25 & 11.31 & 27.42
      & 21.88 & \textbf{41.66} \\
      & Median $R_{\mathrm{flip}}$ (\%)
      & 0.10 & 0.10 & 0.20 & 0.10
      & 0 & 0 & 0 & 0
      & 0 & 0 & 0 & 0
      & 0 & \textbf{0} \\
      & P90 $R_{\mathrm{flip}}$ (\%)
      & 2.50 & 1.90 & 3.40 & 3.70
      & 0.10 & 0.13 & 0.10 & 30.69
      & 0.04 & 0.17 & 0.05 & 0.39
      & 0.78 & \textbf{0.78} \\
    \midrule
    Accuracy drop
      & $>0.5\%$
      & 7.82 & 6.80 & 8.01 & 8.33
      & 0 & 0 & 0 & 13.33
      & 0 & 0 & 0 & 0
      & 9.38 & \textbf{4.42} \\
    \bottomrule
  \end{tabular*}
  \vspace{-12pt}
\end{table*}

\begin{figure}[t]
  \centering
  \includegraphics[width=\columnwidth]{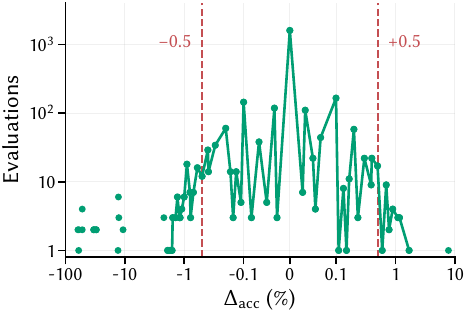}
  \vspace{-24pt}
  \caption{Distribution of $\Delta_{\mathrm{acc}}$.}
  \label{fig:replacement-overall-b}
  \vspace{-20pt}
\end{figure}

To answer RQ3.1, we analyze 2,739 replacement evaluations on kernels
that passed the kernel-level test.
\tabref{tab:replacement-summary} breaks down, by model, how often an
evaluation has any output change or negative flip, how large
$R_{\mathrm{change}}$ and $R_{\mathrm{flip}}$ are, and how often the accuracy drop exceeds 0.5\%.

\figref{fig:replacement-overall-a} shows the results of output change and negative flip.
55.28\% of evaluations have at least one output change, and even
41.66\% have at least one negative flip.
Those changes are usually sparse.
The median $R_{\mathrm{change}}$ is 0.07\% and the median
$R_{\mathrm{flip}}$ is 0\%: on a set of 1,000 model inputs, 0.07\% is
fewer than one model input.
77.8\% of evaluations change at most 1\% of final outputs.
The upper tail is much heavier.
At P90, $R_{\mathrm{change}}$ is 5.92\% and $R_{\mathrm{flip}}$ is
0.78\%. 248 evaluations have $R_{\mathrm{change}}>10\%$, and 24 change outputs for more than half of the model inputs (maximum 94.1\%).

For $\Delta_{\mathrm{acc}}$,
94.7\% of evaluations have $\lvert\Delta_{\mathrm{acc}}\rvert
\le 0.5\%$ (\figref{fig:replacement-overall-b}, between the red dashed lines), and 58.2\% of evaluations have $\Delta_{\mathrm{acc}}=0$.
Outside $\pm 0.5\%$, accuracy drops are more frequent than gains:
121 evaluations (4.42\%) have $\Delta_{\mathrm{acc}}<-0.5\%$,
versus only 24 with $\Delta_{\mathrm{acc}}>0.5\%$.
For $\lvert\Delta_{\mathrm{acc}}\rvert>1\%$, the 64 accuracy drops
(median 12.6\%) are also larger than the 8 accuracy gains (median 1.17\%).
The mean $\Delta_{\mathrm{acc}}$ over all 2,739 evaluations is $-0.44\%$.

\findingbox{Finding 3.1}{\label{sec:replacement:finding-31}
Passing the kernel-level test does not keep model outputs unchanged.
Output changes are common but sparse. $\lvert\Delta_{\mathrm{acc}}\rvert$ usually
stays near zero, and the rare large values are usually drops.}

\subsection{Variation across Configurations and Optimized Kernels}
\label{sec:replacement:rq32}

To answer RQ3.2, we first compare replacement effects across inference configurations. We then compare optimized kernels under the same inference configuration.

Both $R_{\mathrm{change}}$ and $R_{\mathrm{flip}}$ differ sharply
across models (\tabref{tab:replacement-summary}).
Qwen3 and Gemma4 have a heavy tail: the median $R_{\mathrm{change}}$ is
only 0.20\%--0.40\%, while P90 is 13.80\%--33.39\%.
ViT-Base and ViT-Large show output changes in about 60\% of evaluations, typically affecting only a few model inputs per evaluation (P90 near 1\%). ConvNeXT is the opposite of the ViTs: only 25.93\% of evaluations
have $R_{\mathrm{change}}>0$, but P90 is 68.67\%.
For Wav2Vec2, even the median $R_{\mathrm{change}}$ is high at 10.55\%.
For the NLI model, P90 $R_{\mathrm{flip}}$ is only 0.05\%.
Moreover, large accuracy drops are concentrated in a few configurations
(\tabref{tab:replacement-large-drops}):
35 of the 64 evaluations with $\Delta_{\mathrm{acc}}<-1\%$
replace ConvNeXT \texttt{linear} or Gemma \texttt{lm\_head}.
\begin{table}[t]
  \centering
  \caption{Replacement evaluations with $\Delta_{\mathrm{acc}}<-1\%$.}
  \label{tab:replacement-large-drops}
  \small
  \begin{tabular*}{\columnwidth}{@{\extracolsep{\fill}}cccc@{}}
    \toprule
    \textbf{Model} & \textbf{Operator} & \textbf{\# Evals} &
      $\boldsymbol{\#\ \Delta_{\mathrm{acc}}<-1\%}$ \\
    \midrule
    CN & \texttt{linear} & 27 & 18 \\
    G4 & \texttt{lm\_head} & 18 & 8 \\
    G12 & \texttt{lm\_head} & 18 & 9 \\
    Subtotal & & 63 & 35 \\
    All & & 2,739 & 64 \\
    \bottomrule
  \end{tabular*}
  \vspace{-20pt}
\end{table}

At inference, under the same configuration, the choice of optimized kernel can change model behavior substantially, but this does not happen often.
716 of 1,299 configurations have at least two optimized kernels.
Among the 716 configurations with multiple optimized kernels,
the kernels differ by more than 0.5\% in $\Delta_{\mathrm{acc}}$
in 21 configurations, in $R_{\mathrm{change}}$ in 28,
and in $R_{\mathrm{flip}}$ in 15.
In 12 configurations, the difference in $\Delta_{\mathrm{acc}}$ exceeds 1\%.
The gaps in some cases could be larger, we next examine ConvNeXT \texttt{linear} on ImageNet-V2 and Gemma-4-E4B \texttt{lm\_head} on MMLU.

For ConvNeXT \texttt{linear} on ImageNet-V2, the same two of the three optimized kernels produce nearly identical effects across \fpthirtytwo{}, \fpsixteen{}, and \bfsixteen{}.
Their $R_{\mathrm{change}}$ ranges from 68.47\% to 68.67\%, their $R_{\mathrm{flip}}$ ranges from 53.50\% to 53.60\%, and their $\Delta_{\mathrm{acc}}$ is approximately $-52.0\%$. Their mean absolute difference from the reference logits ranges from 0.8035 to 0.8045.
Across all three precisions, the third optimized kernel has $R_{\mathrm{change}}\approx0\%$, $R_{\mathrm{flip}}\approx0\%$, and $\Delta_{\mathrm{acc}}=0$. Its mean absolute difference from the reference logits ranges from approximately $2.44\times10^{-7}$ to $2.90\times10^{-3}$.

To inspect how this difference appears during inference, we added instrumentation to a diagnostic run for one of these evaluations.
At each \texttt{linear} invocation, the instrumentation executes the reference and optimized kernels on the same intermediate input tensor and compares their output tensors.
ConvNeXT invokes the selected \texttt{linear} operator 72 times in every forward pass.
In this run, the reference and optimized output tensors differed at 66 of the 72 invocations.
The first difference appeared at the second linear operation in the first ConvNeXT block, with mean absolute difference $1.54\times10^{-7}$ and maximum absolute difference
$8.1\times10^{-6}$.
At the source-code level, the main visible difference is the assignment of CUDA axes to output indices.
The two optimized kernels associated with the large-drop evaluations use the CUDA $x$ axis for output columns and the $y$ axis for output rows.
The third one reverses these assignments and swaps the grid dimensions accordingly.
This changes which threads process rows and columns and changes the memory-access pattern.
All three kernels nevertheless use the same dot product,
$A[\mathrm{row}\cdot K+k]B[\mathrm{col}\cdot K+k]$, and write the result to the same output location, $C[\mathrm{row}\cdot N+\mathrm{col}]$.

For Gemma-4-E4B \texttt{lm\_head} on MMLU in \fpsixteen{}, two of the three optimized kernels give $R_{\mathrm{change}}=87.2\%$, $R_{\mathrm{flip}}=59.7\%$, and $\Delta_{\mathrm{acc}}=-59.5\%$.
In each of these two evaluations, 840 of the 1,000 model inputs have at least one non-finite option score and an invalid prediction.
The third optimized kernel has $R_{\mathrm{change}}=0$ and $\Delta_{\mathrm{acc}}=0$, with no invalid prediction.

\findingbox{Finding 3.2}{\label{sec:replacement:finding-32}
Replacement effects vary sharply across models, with large accuracy drops concentrated in a few configurations.
Even under the same inference configuration, optimized kernels that pass the kernel-level test can sometimes produce substantially different model behavior.}

\subsection{Characteristics of Affected Model Inputs}
\label{sec:replacement:rq33}

To answer RQ3.3, we analyze 1,863 replacement evaluations
across 10 models for image classification, text classification
(NLI), and multiple-choice QA (MMLU).
Only these task types have per-model-input records of the
reference prediction margin.
For each model input, we obtain the reference class or option scores, apply softmax, and compute the reference prediction margin $m$ as the difference between the top-1 and top-2 probabilities.
We call model inputs with $m<0.05$ low-margin model inputs. We divide replacement evaluations into four groups by $R_{\mathrm{change}}$:
$R_{\mathrm{change}}=0$,
$0<R_{\mathrm{change}}\leq1\%$,
$1\%<R_{\mathrm{change}}\leq10\%$, and
$R_{\mathrm{change}}>10\%$.
\tabref{tab:replacement-margin} groups evaluations by
$R_{\mathrm{change}}$ in its rows and model inputs by reference prediction margin $m$ in its columns.
Each cell reports $r_{\mathrm{change}}$, computed as the number of output changes divided by the number of model inputs in the corresponding group and margin bin.
We count each model input once per evaluation.
The $R_{\mathrm{change}}=0$ group has no output changes by definition, so the following analysis focuses on the three groups with $R_{\mathrm{change}}>0$.

\begin{table}[t]
  \centering
  \caption{$r_{\mathrm{change}}$ (\%) by
    $R_{\mathrm{change}}$ group and $m$.}
  \label{tab:replacement-margin}
  \footnotesize
  \setlength{\tabcolsep}{2pt}
  \begin{tabular*}{\columnwidth}{@{\extracolsep{\fill}}cccccc@{}}
    \toprule
    \multirow{2}{*}{$R_{\mathrm{change}}$} &
      \multirow{2}{*}{\textbf{\# Evals}} &
      \multicolumn{4}{c}{\textbf{Reference prediction margin} $m$} \\
    \cmidrule(l){3-6}
    & & $[0,0.05)$ & $[0.05,0.2)$ & $[0.2,0.5)$ & $[0.5,1]$ \\
    \midrule
    $0$ & 895 & 0 & 0 & 0 & 0 \\
    $(0,1\%]$ & 789 & 2.78 & 0.09 & 0.003 & 0 \\
    $(1\%,10\%]$ & 144 & 8.24 & 0.80 & 0.06 & 0 \\
    $>10\%$ & 35 & 90.23 & 85.20 & 79.39 & 62.36 \\
    \bottomrule
  \end{tabular*}
  \vspace{-24pt}
\end{table}


In the 789 evaluations with $0<R_{\mathrm{change}}\leq1\%$, output changes are concentrated on low-margin model inputs.
This group contains 1,822,000 model inputs in total, including 206,734 with $m<0.05$.
It contains 6,011 output changes in total, of which 5,741 occur on model inputs with $m<0.05$.
Thus, model inputs with $m<0.05$ represent only 11.3\% of the model
inputs counted in this group but account for 95.5\% of its output changes. The same concentration appears in the 144 evaluations with $1\%<R_{\mathrm{change}}\leq10\%$.
This group contains 3,688 output changes, of which 3,212 (87.1\%) occur on model inputs with $m<0.05$.

The pattern changes when output changes become widespread.
In the 35 evaluations with $R_{\mathrm{change}}>10\%$, 26,218 of the 42,045 model inputs in the highest margin bin ($m\geq0.5$) change output, giving $r_{\mathrm{change}}=62.4\%$.
A large reference margin therefore does not guarantee an unchanged output after kernel replacement.

The high-margin result remains after excluding evaluations with invalid predictions.
Five of the 35 evaluations with $R_{\mathrm{change}}>10\%$ include invalid predictions because at least one option score is non-finite for some model inputs.
After excluding these five evaluations, 23,011 of the 38,486 model inputs with $m\geq0.5$ still change output, giving $r_{\mathrm{change}}=59.8\%$.

\findingbox{Finding 3.3}{\label{sec:replacement:finding-33}
Sparse output changes are concentrated on low-margin model inputs. For evaluations with $R_{\mathrm{change}}>10\%$, output changes can also occur for model inputs in the high-margin bin $m\geq0.5$.}

Each replacement evaluation jointly fixes an optimized kernel and an inference configuration, so its results reflect the combined effects of several factors.
\secref{sec:perturbation} uses controlled perturbations to emulate different kernel-level discrepancies.
It helps separate and analyze the effects of different discrepancies and configurations on model behavior.
    
\section{Controlled Kernel Perturbations}
\label{sec:perturbation}

In this section, we use controlled perturbations to examine how kernel-level discrepancy magnitude and inference configuration affect model behavior.  We retain the reference kernel at selected operators in the same 13 models used in the replacement study (\secref{sec:replacement}).
We compare inference with and without perturbations injected into the reference kernel output under the same configuration and on the same model inputs.
We address the following research questions:

\begin{itemize}[labelwidth=!, labelindent=5pt, itemsep=1pt, topsep=1pt, parsep=0pt, partopsep=0pt, leftmargin=15pt]
    \item \textbf{RQ4.1 (Sensitivity to perturbation magnitude).}
    How does model behavior vary with perturbation magnitude
    within a fixed configuration?

    \item \textbf{RQ4.2 (Sensitivity across numerical precisions).}
    How does model behavior differ across numerical precisions at the same perturbation magnitude?

    \item \textbf{RQ4.3 (Sensitivity across operators and datasets).}
    How does model behavior differ across operators and datasets
    at the same perturbation magnitude?

    \item \textbf{RQ4.4 (Empirical relevance to real replacements).}
    Do controlled perturbation measurements provide useful information about
    the effects of real kernel replacements?
\end{itemize}

\para{Perturbation procedure.}
Using the notation in \secref{sec:replacement:procedure},
we add a perturbation $\delta_p$ to the output tensor of the reference kernel 
at every invocation of the selected operator $p$.
The perturbed inference is
\[
F^{\mathrm{pert}}_p(x)
=
F_{>p}\bigl(K^{\mathrm{ref}}_p(F_{<p}(x))+\delta_p\bigr).
\]

For separate modules such as Linear, RMSNorm, or GELU, forward hooks
perturb their output tensors. For direct function calls, a wrapper calls the original function and adds the perturbation to the returned tensor. 
For suboperations within Attention, such as Softmax or the QK matmul, do not expose separate module boundaries. We therefore modify the attention function to add the perturbation immediately after the selected suboperation, before its output is used by subsequent attention computations.

For each output element $j$, we set
$\delta_{p,j}=s_j a_j$,
where $a_j$ is sampled uniformly from $[\varepsilon/10,\varepsilon]$ and $s_j$ is $-1$ or $+1$ with equal probability.
Here, $\varepsilon$ is the upper bound on the sampled per-element perturbation magnitude. We use $\varepsilon$ to describe perturbation magnitude in the following analysis because the lower bound is fixed at $\varepsilon/10$, varying $\varepsilon$ scales the entire magnitude distribution.
We independently sample the magnitude and sign for each output element at every invocation.
When the reference kernel produces an \fpsixteen{} or \bfsixteen{} output tensor, we sample perturbation magnitudes in \fpthirtytwo{}, apply the signs, and convert the perturbations to the same data type as the output tensor before adding them to it.
The reference and perturbed inferences share identical configurations. For \fpthirtytwo{}, we use upper bounds $\varepsilon \in \{10^{-7},10^{-6},10^{-5},10^{-4}\}$. For \fpsixteen{} and \bfsixteen{}, we additionally include $\varepsilon=10^{-3}$ and $10^{-2}$.

\para{Example.}
In the perturbation path of \figref{fig:replacement-example}, input
$X$ passes through the PyTorch ReLU kernel in panel (a), producing
$Y_{\mathrm{ref}}$. We pass $Y_{\mathrm{pert}}=Y_{\mathrm{ref}}+\delta_p$
to global average pooling and the remaining ConvNeXT computation,
repeating this procedure at every invocation of the selected operator.

\subsection{Sensitivity to Perturbation Magnitude}
\label{sec:perturbation:rq41}

\begin{figure*}[t]
  \centering
  \includegraphics[width=\textwidth]{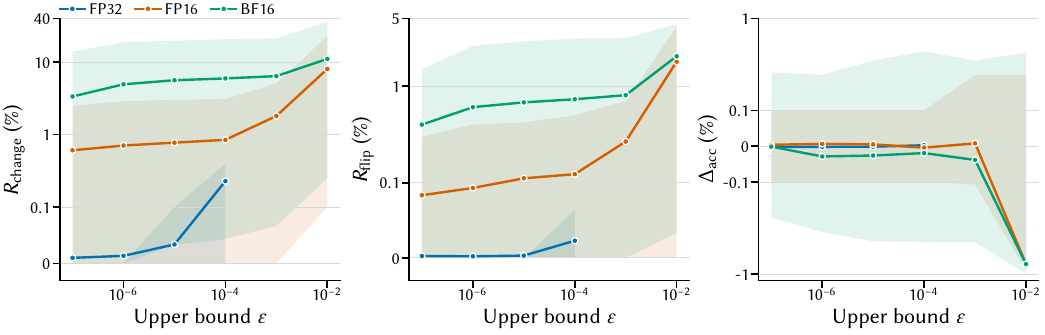}
  \vspace{-25pt}
  \caption{$R_{\mathrm{change}}$, $R_{\mathrm{flip}}$, and $\Delta_{\mathrm{acc}}$ as $\varepsilon$ varies.}
  \label{fig:perturbation-response}
  \vspace{-16pt}
\end{figure*}

We vary $\varepsilon$ within each fixed configuration.
Figure~\ref{fig:perturbation-response} shows the mean
$R_{\mathrm{change}}$, $R_{\mathrm{flip}}$, and $\Delta_{\mathrm{acc}}$
across configurations for each precision and value of $\varepsilon$,
with shading for the 10th--90th percentile range.

The mean $R_{\mathrm{change}}$ and $R_{\mathrm{flip}}$ generally increase with $\varepsilon$. In 996 of the 1,299 configurations, $R_{\mathrm{change}}$ is higher at the largest tested value of $\varepsilon$ than at the smallest.

However, in some configurations, $R_{\mathrm{change}}$ does not increase monotonically as $\varepsilon$ increases. For example, for Wav2Vec2 on LibriSpeech test-other with \bfsixteen{} and \texttt{conv1d}, $R_{\mathrm{change}}$ decreases from 33.98\% to 28.52\% as $\varepsilon$ increases from $10^{-7}$ to $10^{-6}$.
In 670 of the 1,299 configurations, $R_{\mathrm{change}}$ decreases at least once as $\varepsilon$ increases. In 60 of these configurations, at least one such decrease exceeds 1\%.

\findingbox{Finding 4.1}{
Larger $\varepsilon$ values generally produce larger
$R_{\mathrm{change}}$ and $R_{\mathrm{flip}}$ values across configurations,
but individual response curves are not uniformly monotone.
}

\subsection{Sensitivity across Numerical Precisions}
\label{sec:perturbation:rq42}

\begin{figure}[t]
  \centering
  \includegraphics[width=\columnwidth]{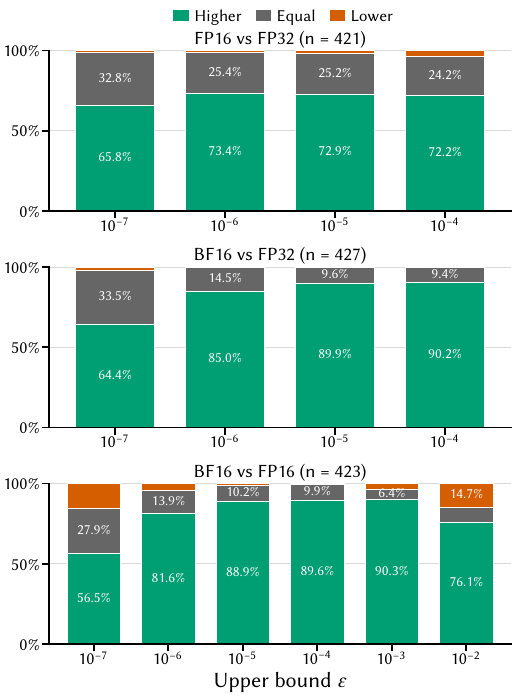}
  \vspace{-24pt}
  \caption{Pairwise comparisons of $R_{\mathrm{change}}$ across numerical precisions.}
  \label{fig:perturbation-precision-pairs}
  \vspace{-20pt}
\end{figure}

We next compare $R_{\mathrm{change}}$ across numerical precisions while holding the model, operator, dataset, and $\varepsilon$ fixed.
Figure~\ref{fig:perturbation-precision-pairs} compares \fpsixteen{}--\fpthirtytwo{}, \bfsixteen{}--\fpthirtytwo{}, and \bfsixteen{}--\fpsixteen{} at the values of $\varepsilon$ shared by each pair.
Each bar shows the percentages of these comparisons in which the first precision's $R_{\mathrm{change}}$ is higher than, equal to, or lower than that of the second precision.

At the same $\varepsilon$, $R_{\mathrm{change}}$ is usually
higher for \fpsixteen{} and \bfsixteen{} than for \fpthirtytwo{}.
Across the four shared values of $\varepsilon$, the fraction
of configurations with a higher $R_{\mathrm{change}}$ for \fpsixteen{}
than for \fpthirtytwo{} remains between 65.8\% and 73.4\%.
For \bfsixteen{}, this fraction increases from 64.4\% at
$\varepsilon=10^{-7}$ to 90.2\% at $\varepsilon=10^{-4}$.

At each of the six shared values of $\varepsilon$, $R_{\mathrm{change}}$ is higher for \bfsixteen{} than for \fpsixteen{} in more than half of the 423 matched configurations.
This percentage increases from 56.5\% at $\varepsilon=10^{-7}$ to 90.3\% at $10^{-3}$, then decreases to 76.1\% at $10^{-2}$. However, at $\varepsilon=10^{-2}$, $R_{\mathrm{change}}$ is lower for \bfsixteen{} than for \fpsixteen{} in 62 of these configurations.
The ranking can also change as $\varepsilon$ increases.
For example, for Qwen3-8B on GSM8K with \texttt{attention\_softmax}, $R_{\mathrm{change}}$ at $\varepsilon=10^{-3}$ is higher for \bfsixteen{} than for \fpsixteen{}: 22.7\% versus 15.4\%.
At $\varepsilon=10^{-2}$, it is lower for \bfsixteen{} than for \fpsixteen{}: 59.4\% versus 65.2\%.

\findingbox{Finding 4.2}{
Sensitivity to perturbations differs across numerical precisions and these differences vary with the configuration and perturbation magnitude.
}

\subsection{Sensitivity across Operators and Datasets}
\label{sec:perturbation:rq43}

\begin{figure*}[t]
  \centering
  \includegraphics[width=\textwidth]{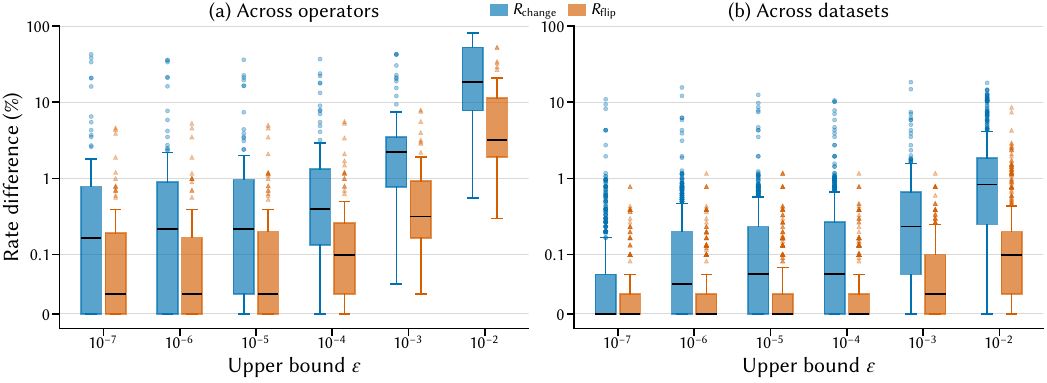}
  \vspace{-24pt}
  \caption{Variation in $R_{\mathrm{change}}$ and $R_{\mathrm{flip}}$ across operators and datasets.}
  \label{fig:perturbation-operator-dataset}
  \vspace{-16pt}
\end{figure*}

Figure~\ref{fig:perturbation-operator-dataset} summarizes differences in $R_{\mathrm{change}}$ and $R_{\mathrm{flip}}$ across operators and datasets.
In the left panel, we fix the model, precision, dataset, and $\varepsilon$, and compare the measured operators. For each metric, we subtract the lowest value from the highest value.
In the right panel, we fix the model, precision, operator, and $\varepsilon$, and compare datasets for the same task, such as ImageNet-V2 and ImageNet-A for image classification.
For each metric, we calculate the absolute difference between its values on the two datasets. Boxes show the 25th to 75th percentiles, and black lines mark the medians.

At $\varepsilon=10^{-2}$, the median difference between the highest and lowest operator rates is 18.4\% for $R_{\mathrm{change}}$ and 3.23\% for $R_{\mathrm{flip}}$. When only the dataset changes,
the median absolute differences are 0.83\% and 0.10\% respectively.
Some comparisons show substantially larger differences.

For example, on Gemma-4 12B with \fpsixteen{} and MMLU at
$\varepsilon=10^{-2}$, $R_{\mathrm{change}}$ is
73.6\% for \texttt{post\_attention\_rmsnorm},
26.3\% for \texttt{attention\_softmax},
3.7\% for \texttt{linear},
and 0.0\% for \texttt{lm\_head}.
The corresponding $R_{\mathrm{flip}}$ values are 53.0\%, 12.8\%, 1.9\%, and 0.0\%,
while $\Delta_{\mathrm{acc}}$ is $-47.0\%$, $-7.9\%$, $-1.2\%$, and 0.0\%.
Thus, even at the same $\varepsilon$, the choice of operator can sharply affect $R_{\mathrm{change}}$, $R_{\mathrm{flip}}$ and $\Delta_{\mathrm{acc}}$.

Changing the dataset also shows that the three metrics can sometimes change in different directions.
For ViT-Base with \fpsixteen{} \texttt{attention\_softmax} at $\varepsilon=10^{-2}$, $R_{\mathrm{change}}$ is 6.27\% on ImageNet-V2 and 18.57\% on ImageNet-A, nearly three times as high, while $R_{\mathrm{flip}}$ remains closer at 2.03\% and 1.27\%.
$\Delta_{\mathrm{acc}}$ even differs in sign: $-0.63\%$ and $+0.40\%$, respectively.

\findingbox{Finding 4.3}{
The same $\varepsilon$ can have different effects on model behavior across operators and datasets. Sensitivity observed for one operator or dataset therefore might not hold for another.
}

\subsection{Empirical Relevance to Real Replacements}
\label{sec:perturbation:rq44}

Unlike independently sampled perturbations, real kernel-level discrepancies
may depend on the input tensor and kernel implementation and follow
consistent patterns across output elements. We use a simple predictor to
assess whether the perturbation results provide useful for real replacement effects,
without claiming a practical predictor for deployment.

KIP is a random forest trained on 6,928 perturbation evaluations to predict
$R_{\mathrm{change}}$ and $R_{\mathrm{flip}}$. We evaluate it on 2,739 real
replacement evaluations, using each replacement's measured maximum
absolute kernel-level discrepancy as an input.
A budget pair specifies an upper limit for each of $R_{\mathrm{change}}$ and $R_{\mathrm{flip}}$. For example, $(2\%,1\%)$ sets the two limits to 2\% and 1\%, respectively. KIP flags a replacement if either predicted rate exceeds its limit. A replacement exceeds the budget pair if its observed $R_{\mathrm{change}}$ or $R_{\mathrm{flip}}$ exceeds the respective limit.

\begin{figure}[t]
  \centering
  \includegraphics[width=\columnwidth]{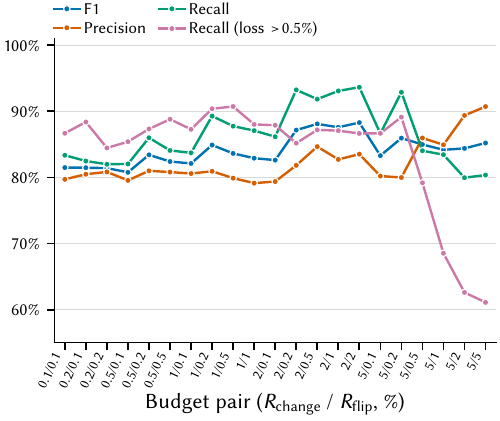}
  \vspace{-24pt}
  \caption{KIP prediction performance across 21 budget pairs.}
  \label{fig:kip-budget-path}
  \vspace{-20pt}
\end{figure}

We evaluate 21 budget pairs. For each pair, we report the mean over five balanced repetitions, each containing 1,000 sampled replacement evaluations: 500 that exceed either budget and 500 that exceed neither.
Figure~\ref{fig:kip-budget-path} summarizes KIP's prediction performance. Across the 21 pairs, F1 ranges from 80.78\% to 88.29\%, precision from 79.14\% to 90.71\%, and recall from 79.96\% to 93.64\%.
Among real replacement evaluations with an observed accuracy drop above 0.5\%, KIP identifies 61.10\%--90.73\% as exceeding at least one budget. These results show that controlled perturbation measurements provide useful information about real replacement effects within the evaluated configurations.

\findingbox{Finding 4.4}{
Controlled perturbations and real kernel replacements show related effects on model behavior within the evaluated configurations.
}

    \section{Discussion}
\label{sec:discussion}


\subsection{Implications for Kernel Evaluation}
\label{sec:discussion:implications}

\para{Practical relevance of controlled perturbations.}
Controlled perturbations characterize how model behavior responds to perturbation magnitude, with responses varying across numerical precisions, operators, and datasets (\secref{sec:perturbation}).
The transfer analysis in \secref{sec:perturbation:rq44} provides evidence that these measurements also contain useful information about real replacement effects within the covered configurations.
This supports using perturbation measurements as an empirical reference for assessing how kernel-level discrepancies affect model behavior in the target configuration.

\para{Potential for advanced assessment.}
Given an optimized kernel's measured maximum absolute kernel-level discrepancy and its configuration, KIP uses perturbation measurements to estimate output-change and negative-flip rates.
Before full model inference, these estimates may help prioritize optimized kernels according to application requirements.
Whether this reduces validation cost while retaining acceptable model behavior requires prospective evaluation within an optimization workflow.

\para{Tolerance threshold selection.}
Passing the kernel-level test does not ensure unchanged model outputs or accuracy.
Tolerance choices should therefore be assessed through output-change rates, negative-flip rates, and accuracy changes in the target configuration, with application requirements defining acceptable model behavior.
Our study does not establish a universally safe threshold or how much existing thresholds should be adjusted. We do not directly evaluate how changing tolerance thresholds
affects model behavior.

\subsection{Threats to Validity}
\label{sec:discussion:validity}

\para{Experimental coverage.}
Generality is limited by kernel sources, model coverage, model-input
subsets, and the single GPU architecture and software stack.
Evaluations reuse configurations and model inputs, so they are not
independent deployment trials. Reported frequencies also depend on
how many accepted kernels each configuration contributes.

\para{Perturbation model and measurements.}
Real kernel errors can be biased, correlated, and input-dependent.
A perturbation bound and a measured maximum absolute discrepancy therefore need not imply equivalent numerical effects, and equal absolute bounds need not represent equal relative errors or total perturbations across configurations.
A limited number of model inputs, a discrete magnitude grid, and one perturbation seed limit conclusions about unobserved changes and non-monotonic responses.
Precision comparisons require common magnitude ranges.

\para{Prediction scope and selection.}
KIP is trained on perturbations and evaluated on replacements within the same 1,299 configurations, so generalization to unseen models or configurations remains unestablished.
Predictor and evaluation choices were explored with replacement results available, including retrospective selection of the resampling seed.
Metrics are conditional on these choices, balanced sampling, and configuration weighting.
The estimates do not replace the final model-level evaluation or guarantee accuracy retention on future model inputs.

    \section{Related Work}
\label{sec:related-work}

\para{Benchmarking kernel generation.}
Existing benchmarks mainly evaluate generated kernels through differential testing and performance measurement. KernelBench~\cite{ouyang2025kernelbench} establishes this basic protocol, while TritonBench~\cite{li2025tritonbench} and MultiKernelBench~\cite{wen2025multikernelbench} extend it to Triton operators and heterogeneous accelerators. More recent benchmarks move toward realistic deployment settings, including inference workloads in FlashInfer-Bench~\cite{xing2026flashinferbench}, integration failures in FastKernels~\cite{oliaro2026fastkernels}, and unseen configurations in AgentKernelArena~\cite{younesian2026agentkernelarena}. 
These efforts broaden where kernels are tested, but their correctness criteria remain largely local and do not establish unchanged model behavior after integration.

\para{Agentic kernel generation.}
Recent systems increasingly automate kernel optimization through learning, feedback, and domain specialization. CUDA-L1, CUDA Agent, KernelBlaster, and CudaForge improve generation with reinforcement learning, accumulated optimization experience, or hardware feedback~\cite{li2026cudal1,dai2026cudaagent,dong2026kernelblaster,zhang2025cudaforge}. CuTeGen, CuBridge, and CUDAMaster target specialized kernel domains~\cite{saba2026cutegen,ma2026cubridge,han2026multiscenario}, while KLineage and CAKE explore reusable optimization knowledge and compiler-agent co-design~\cite{zhang2026klineage,ye2026cake}. 
These systems primarily optimize kernel-level performance subject to a kernel-level test; our work instead studies model behavior after such kernels are accepted.

\para{Kernel correctness validation.}
Prior work strengthens kernel correctness from several directions, including explicit numerical contracts~\cite{veit2026kernelcontracts,xu2026traininginferencecontracts}, fuzzing and stronger equivalence or safety checks~\cite{sarkar2026illusion,chatterjee2025proofwright,he2026model2kernel,driscoll2025equivalence}, and roundoff-error analysis~\cite{solovyev2018fptaylor,darulova2018daisy,daumas2010gappa,moscato2017precisa}. Other studies show that numerical differences can affect DNN outputs and depend on execution context~\cite{sharif2019approxhpvm,jiang2023arbitor,badash2026hawkeye,yuan2025nondeterminism,pape2026silent,zhu2026heal,chen2026integeralibi}. Tolerance-based conformance work further shows that local passing does not guarantee interchangeability~\cite{chen2026toleranceconformance}. 
We build on these observations by measuring how kernel-level discrepancies in real generated kernels translate into model-input-level output changes and negative flips.

    \section{Conclusion}
\label{sec:conclusion}

Generated GPU kernels are admitted today by a tolerance threshold that
do not establish how its numerical discrepancies affect model behavior.
Through real kernel replacements and controlled perturbations across
13 models, we examine this gap.
Our replacement evaluations show that output changes and negative
flips can occur even when accuracy changes little, and some replacements
cause large accuracy drops.
Different optimized kernels that pass the test can also produce
substantially different effects under the same inference configuration.
Our perturbation evaluations show that these effects depend on
numerical precision, operator, and dataset, and that increasing
perturbation magnitude does not always increase output changes.
Within the evaluated configurations, controlled perturbation results could
be related the effects of real kernel replacements. These findings motivate correctness criteria for kernel benchmarks and
optimization agents that account for model behavior in the target
inference configuration and the application's requirements.

    \bibliographystyle{mlsys2025}
    \bibliography{bib/references}

    \clearpage
    \appendix
\setcounter{table}{0}
\setcounter{figure}{0}
\renewcommand{\thetable}{\thesection.\arabic{table}}
\renewcommand{\thefigure}{\thesection.\arabic{figure}}
\renewcommand{\theHtable}{supplement.\thesection.\arabic{table}}
\renewcommand{\theHfigure}{supplement.\thesection.\arabic{figure}}

\section{Additional Real Kernel Replacement Results}
\label{app:replacement-results}

Tables~\ref{app:replacement-summary:1} and~\ref{app:replacement-summary:2}
summarize all 2,739 replacement evaluations by model, numerical precision,
and dataset. Each evaluation compares reference inference with inference
using one optimized kernel on the same model inputs. Within each table
row, every replacement evaluation receives equal weight.

\para{Measurements \& metrics.}
An output change occurs when the final output after replacement differs
from the reference output on the same model input. We compare the selected
option on MMLU, the parsed numerical answer on GSM8K, the top-1 label on
ImageNet-V2/A/Sketch, the extracted span on SQuAD~v2 and AdvQA, the
predicted label on NLI, and the decoded transcript on LibriSpeech.
A negative flip occurs when the reference output is correct and the output
after replacement is not. Correctness requires an exact match to the ground
truth. The output-change rate, $R_{\mathrm{change}}$, and negative-flip
rate, $R_{\mathrm{flip}}$, divide their respective event counts by the total
number of model inputs in the evaluation.
Accuracy change, $\Delta_{\mathrm{acc}}$, is accuracy after replacement minus
reference accuracy; a negative value indicates an accuracy drop.
We report both rates and accuracy changes in \%.

\para{Table statistics.}
The median and P90 summarize the distribution across replacement evaluations
within each row. Quantiles use linear interpolation. Min and Max give the
smallest and largest accuracy changes in the group. $E$ is the number of
replacement evaluations, and $D_{>1}$ counts evaluations with accuracy drops
greater than 1\%. Values are rounded to two decimal places, with exact zeros
shown as 0.

\para{Models \& datasets.}
The tables include every evaluated combination of model, numerical precision,
and dataset. DeBERTa-v3-L (QA) and DeBERTa-v3-L (NLI) denote the extractive
QA and NLI models, respectively. MNLI uses the matched validation subset;
ANLI combines R1, R2, and R3. LibriSpeech clean and other denote the
test-clean and test-other subsets. Results for individual optimized kernels
are retained in the accompanying data.

\begin{table*}[!tp]
\centering
\caption{Real kernel replacement results by model, numerical precision, and dataset (1/2). Rates and accuracy changes are in \%. $E$ counts replacement evaluations; $D_{>1}$ counts evaluations with accuracy drops greater than 1\%. Median, P90, Min, and Max summarize evaluations within each row.}
\label{app:replacement-summary:1}
\fontsize{8}{10}\selectfont
\setlength{\tabcolsep}{2pt}
\renewcommand{\arraystretch}{1.0}
\begin{tabular*}{\textwidth}{@{\extracolsep{\fill}}cccccccccccc@{}}
\toprule
\multirow{2}{*}{Model} & \multirow{2}{*}{Dataset} & \multirow{2}{*}{Precision} & \multirow{2}{*}{$E$} & \multicolumn{2}{c}{$R_{\mathrm{change}}$} & \multicolumn{2}{c}{$R_{\mathrm{flip}}$} & \multicolumn{3}{c}{$\Delta_{\mathrm{acc}}$} & \multirow{2}{*}{$D_{>1}$} \\
\cmidrule(lr){5-6}\cmidrule(lr){7-8}\cmidrule(lr){9-11}
 & & & & Median & P90 & Median & P90 & Min & Median & Max & \\
\midrule
\multirow{6}{*}{Qwen3-8B} & \multirow{3}{*}{MMLU} & \fpthirtytwo{} & 49 & 0 & 0 & 0 & 0 & 0 & 0 & 0 & 0 \\
 &  & \fpsixteen{} & 49 & 0.10 & 0.30 & 0 & 0.10 & $-0.10$ & 0.10 & 0.20 & 0 \\
 &  & \bfsixteen{} & 49 & 0.90 & 1.30 & 0.20 & 0.50 & $-0.20$ & 0.10 & 0.30 & 0 \\
\arrayrulecolor{black!30}\cline{2-12}\arrayrulecolor{black}
 & \multirow{3}{*}{GSM8K} & \fpthirtytwo{} & 49 & 0 & 0 & 0 & 0 & 0 & 0 & 0 & 0 \\
 &  & \fpsixteen{} & 49 & 2.40 & 3.20 & 0.30 & 0.50 & $-0.50$ & 0 & 0.20 & 0 \\
 &  & \bfsixteen{} & 49 & 18.00 & 20.10 & 2.80 & 3.20 & $-1.70$ & $-0.20$ & 0.40 & 9 \\
\midrule[0.5pt]
\multirow{6}{*}{Qwen3-14B} & \multirow{3}{*}{MMLU} & \fpthirtytwo{} & 49 & 0 & 0 & 0 & 0 & 0 & 0 & 0 & 0 \\
 &  & \fpsixteen{} & 49 & 0.10 & 0.40 & 0 & 0.20 & $-0.10$ & 0 & 0.20 & 0 \\
 &  & \bfsixteen{} & 49 & 1.00 & 1.40 & 0.30 & 0.50 & $-0.40$ & 0 & 0.40 & 0 \\
\arrayrulecolor{black!30}\cline{2-12}\arrayrulecolor{black}
 & \multirow{3}{*}{GSM8K} & \fpthirtytwo{} & 49 & 0 & 0 & 0 & 0 & 0 & 0 & 0 & 0 \\
 &  & \fpsixteen{} & 49 & 1.90 & 2.70 & 0.20 & 0.40 & $-0.50$ & $-0.20$ & 0.30 & 0 \\
 &  & \bfsixteen{} & 49 & 14.10 & 15.32 & 1.90 & 2.50 & $-1.80$ & $-0.20$ & 0.70 & 1 \\
\midrule[0.5pt]
\multirow{6}{*}{Gemma-4-E4B} & \multirow{3}{*}{MMLU} & \fpthirtytwo{} & 52 & 0 & 0 & 0 & 0 & $-13.00$ & 0 & 0 & 3 \\
 &  & \fpsixteen{} & 52 & 0.30 & 0.50 & 0.10 & 0.20 & $-59.50$ & 0 & 0.20 & 2 \\
 &  & \bfsixteen{} & 52 & 1.50 & 2.00 & 0.60 & 1.00 & $-61.50$ & $-0.10$ & 0.70 & 3 \\
\arrayrulecolor{black!30}\cline{2-12}\arrayrulecolor{black}
 & \multirow{3}{*}{GSM8K} & \fpthirtytwo{} & 52 & 0 & 0 & 0 & 0 & 0 & 0 & 0 & 0 \\
 &  & \fpsixteen{} & 52 & 5.90 & 6.89 & 0.80 & 1.09 & $-0.90$ & 0 & 0.50 & 0 \\
 &  & \bfsixteen{} & 52 & 33.45 & 35.00 & 3.40 & 4.40 & $-2.20$ & $-0.05$ & 1.70 & 7 \\
\midrule[0.5pt]
\multirow{6}{*}{Gemma-4-12B} & \multirow{3}{*}{MMLU} & \fpthirtytwo{} & 52 & 0 & 0 & 0 & 0 & $-12.80$ & 0 & 0 & 3 \\
 &  & \fpsixteen{} & 52 & 0.10 & 0.20 & 0 & 0.10 & $-12.80$ & 0 & 0.20 & 3 \\
 &  & \bfsixteen{} & 52 & 0.90 & 1.20 & 0.30 & 0.49 & $-12.60$ & 0.10 & 0.50 & 3 \\
\arrayrulecolor{black!30}\cline{2-12}\arrayrulecolor{black}
 & \multirow{3}{*}{GSM8K} & \fpthirtytwo{} & 52 & 0 & 0.10 & 0 & 0.10 & $-0.10$ & 0 & 0.10 & 0 \\
 &  & \fpsixteen{} & 52 & 3.20 & 3.89 & 0.55 & 0.90 & $-0.90$ & 0 & 0.70 & 0 \\
 &  & \bfsixteen{} & 52 & 20.00 & 21.00 & 3.70 & 4.20 & $-1.40$ & $-0.30$ & 0.90 & 6 \\
\midrule[0.5pt]
\multirow{9}{*}{ViT-Base} & \multirow{3}{*}{ImageNet-V2} & \fpthirtytwo{} & 42 & 0 & 0 & 0 & 0 & 0 & 0 & 0 & 0 \\
 &  & \fpsixteen{} & 45 & 0.03 & 0.03 & 0 & 0.03 & $-0.03$ & 0 & 0.03 & 0 \\
 &  & \bfsixteen{} & 45 & 0.23 & 0.33 & 0.03 & 0.07 & $-0.17$ & 0 & 0.10 & 0 \\
\arrayrulecolor{black!30}\cline{2-12}\arrayrulecolor{black}
 & \multirow{3}{*}{ImageNet-A} & \fpthirtytwo{} & 42 & 0 & 0 & 0 & 0 & 0 & 0 & 0 & 0 \\
 &  & \fpsixteen{} & 45 & 0.10 & 0.13 & 0 & 0.03 & $-0.03$ & 0 & 0.03 & 0 \\
 &  & \bfsixteen{} & 45 & 0.83 & 1.10 & 0.10 & 0.17 & $-0.13$ & 0 & 0.17 & 0 \\
\arrayrulecolor{black!30}\cline{2-12}\arrayrulecolor{black}
 & \multirow{3}{*}{ImageNet-Sketch} & \fpthirtytwo{} & 42 & 0 & 0 & 0 & 0 & 0 & 0 & 0 & 0 \\
 &  & \fpsixteen{} & 45 & 0.07 & 0.10 & 0 & 0.03 & $-0.03$ & 0 & 0.07 & 0 \\
 &  & \bfsixteen{} & 45 & 0.87 & 1.00 & 0.07 & 0.17 & $-0.17$ & 0 & 0.17 & 0 \\
\midrule[0.5pt]
\multirow{9}{*}{ViT-Large} & \multirow{3}{*}{ImageNet-V2} & \fpthirtytwo{} & 42 & 0 & 0 & 0 & 0 & 0 & 0 & 0 & 0 \\
 &  & \fpsixteen{} & 45 & 0.03 & 0.10 & 0 & 0.03 & $-0.03$ & 0 & 0.13 & 0 \\
 &  & \bfsixteen{} & 45 & 0.30 & 0.43 & 0.07 & 0.13 & $-0.17$ & 0.03 & 0.30 & 0 \\
\arrayrulecolor{black!30}\cline{2-12}\arrayrulecolor{black}
 & \multirow{3}{*}{ImageNet-A} & \fpthirtytwo{} & 42 & 0 & 0 & 0 & 0 & $-0.07$ & 0 & 0 & 0 \\
 &  & \fpsixteen{} & 45 & 0.17 & 0.27 & 0.03 & 0.05 & $-0.13$ & $-0.03$ & 0.03 & 0 \\
 &  & \bfsixteen{} & 45 & 1.17 & 1.40 & 0.10 & 0.23 & $-0.20$ & 0 & 0.10 & 0 \\
\arrayrulecolor{black!30}\cline{2-12}\arrayrulecolor{black}
 & \multirow{3}{*}{ImageNet-Sketch} & \fpthirtytwo{} & 42 & 0 & 0 & 0 & 0 & 0 & 0 & 0 & 0 \\
 &  & \fpsixteen{} & 45 & 0.10 & 0.20 & 0 & 0.03 & $-0.03$ & 0 & 0.03 & 0 \\
 &  & \bfsixteen{} & 45 & 0.77 & 0.97 & 0.07 & 0.13 & $-0.07$ & 0 & 0.17 & 0 \\
\midrule[0.5pt]
\multirow{9}{*}{DINOv2-Large} & \multirow{3}{*}{ImageNet-V2} & \fpthirtytwo{} & 9 & 0 & 0 & 0 & 0 & 0 & 0 & 0 & 0 \\
 &  & \fpsixteen{} & 9 & 0 & 0.03 & 0 & 0.03 & $-0.03$ & 0 & 0 & 0 \\
 &  & \bfsixteen{} & 6 & 0.07 & 0.23 & 0.02 & 0.03 & 0 & 0.02 & 0.13 & 0 \\
\arrayrulecolor{black!30}\cline{2-12}\arrayrulecolor{black}
 & \multirow{3}{*}{ImageNet-A} & \fpthirtytwo{} & 9 & 0 & 0 & 0 & 0 & 0 & 0 & 0 & 0 \\
 &  & \fpsixteen{} & 9 & 0.20 & 0.27 & 0.03 & 0.10 & $-0.07$ & 0 & 0.07 & 0 \\
 &  & \bfsixteen{} & 6 & 0.43 & 1.03 & 0.12 & 0.25 & $-0.10$ & 0 & 0 & 0 \\
\arrayrulecolor{black!30}\cline{2-12}\arrayrulecolor{black}
 & \multirow{3}{*}{ImageNet-Sketch} & \fpthirtytwo{} & 9 & 0 & 0 & 0 & 0 & 0 & 0 & 0 & 0 \\
 &  & \fpsixteen{} & 9 & 0.17 & 0.20 & 0.07 & 0.07 & $-0.10$ & 0 & 0.03 & 0 \\
 &  & \bfsixteen{} & 6 & 0.25 & 0.60 & 0.05 & 0.15 & $-0.07$ & 0 & 0.07 & 0 \\
\bottomrule
\end{tabular*}
\end{table*}

\begin{table*}[!tp]
\centering
\caption{Real kernel replacement results by model, numerical precision, and dataset (2/2). Rates and accuracy changes are in \%. $E$ counts replacement evaluations; $D_{>1}$ counts evaluations with accuracy drops greater than 1\%. Median, P90, Min, and Max summarize evaluations within each row.}
\label{app:replacement-summary:2}
\fontsize{8}{10}\selectfont
\setlength{\tabcolsep}{2pt}
\renewcommand{\arraystretch}{1.0}
\begin{tabular*}{\textwidth}{@{\extracolsep{\fill}}cccccccccccc@{}}
\toprule
\multirow{2}{*}{Model} & \multirow{2}{*}{Dataset} & \multirow{2}{*}{Precision} & \multirow{2}{*}{$E$} & \multicolumn{2}{c}{$R_{\mathrm{change}}$} & \multicolumn{2}{c}{$R_{\mathrm{flip}}$} & \multicolumn{3}{c}{$\Delta_{\mathrm{acc}}$} & \multirow{2}{*}{$D_{>1}$} \\
\cmidrule(lr){5-6}\cmidrule(lr){7-8}\cmidrule(lr){9-11}
 & & & & Median & P90 & Median & P90 & Min & Median & Max & \\
\midrule
\multirow{9}{*}{ConvNeXT-Base} & \multirow{3}{*}{ImageNet-V2} & \fpthirtytwo{} & 15 & 0 & 41.20 & 0 & 32.16 & $-52.07$ & 0 & 0 & 2 \\
 &  & \fpsixteen{} & 15 & 0 & 41.20 & 0 & 32.14 & $-52.03$ & 0 & 0 & 2 \\
 &  & \bfsixteen{} & 15 & 0 & 41.15 & 0 & 32.13 & $-52.03$ & 0 & 0 & 2 \\
\arrayrulecolor{black!30}\cline{2-12}\arrayrulecolor{black}
 & \multirow{3}{*}{ImageNet-A} & \fpthirtytwo{} & 15 & 0 & 56.40 & 0 & 18.40 & $-30.37$ & 0 & 0 & 2 \\
 &  & \fpsixteen{} & 15 & 0 & 56.43 & 0 & 18.43 & $-30.40$ & 0 & 0 & 2 \\
 &  & \bfsixteen{} & 15 & 0 & 56.67 & 0 & 18.39 & $-30.20$ & 0 & 0 & 2 \\
\arrayrulecolor{black!30}\cline{2-12}\arrayrulecolor{black}
 & \multirow{3}{*}{ImageNet-Sketch} & \fpthirtytwo{} & 15 & 0 & 47.82 & 0 & 20.94 & $-33.30$ & 0 & 0 & 2 \\
 &  & \fpsixteen{} & 15 & 0 & 47.88 & 0 & 20.90 & $-33.23$ & 0 & 0.07 & 2 \\
 &  & \bfsixteen{} & 15 & 0 & 47.95 & 0 & 20.93 & $-33.27$ & 0 & 0 & 2 \\
\midrule[0.5pt]
\multirow{9}{*}{CLIP ViT-B/32} & \multirow{3}{*}{ImageNet-V2} & \fpthirtytwo{} & 11 & 0 & 0 & 0 & 0 & 0 & 0 & 0 & 0 \\
 &  & \fpsixteen{} & 11 & 0.13 & 0.20 & 0 & 0 & 0 & 0 & 0.03 & 0 \\
 &  & \bfsixteen{} & 8 & 0.47 & 1.13 & 0 & 0.32 & $-0.17$ & 0 & 0 & 0 \\
\arrayrulecolor{black!30}\cline{2-12}\arrayrulecolor{black}
 & \multirow{3}{*}{ImageNet-A} & \fpthirtytwo{} & 11 & 0 & 0 & 0 & 0 & 0 & 0 & 0 & 0 \\
 &  & \fpsixteen{} & 11 & 0.27 & 0.30 & 0 & 0.03 & $-0.03$ & 0 & 0 & 0 \\
 &  & \bfsixteen{} & 8 & 0.67 & 1.57 & 0 & 0.12 & $-0.07$ & 0 & 0 & 0 \\
\arrayrulecolor{black!30}\cline{2-12}\arrayrulecolor{black}
 & \multirow{3}{*}{ImageNet-Sketch} & \fpthirtytwo{} & 11 & 0 & 0 & 0 & 0 & 0 & 0 & 0 & 0 \\
 &  & \fpsixteen{} & 11 & 0.23 & 0.40 & 0 & 0.07 & $-0.03$ & 0 & 0.07 & 0 \\
 &  & \bfsixteen{} & 8 & 0.67 & 1.77 & 0.02 & 0.22 & $-0.10$ & 0 & 0.13 & 0 \\
\midrule[0.5pt]
\multirow{6}{*}{DeBERTa-v3-L (QA)} & \multirow{3}{*}{SQuAD v2} & \fpthirtytwo{} & 18 & 0 & 0 & 0 & 0 & 0 & 0 & 0 & 0 \\
 &  & \fpsixteen{} & 19 & 0.03 & 0.06 & 0 & 0.03 & $-0.03$ & 0 & 0.11 & 0 \\
 &  & \bfsixteen{} & 19 & 0.33 & 0.48 & 0.11 & 0.22 & $-0.11$ & 0 & 0.06 & 0 \\
\arrayrulecolor{black!30}\cline{2-12}\arrayrulecolor{black}
 & \multirow{3}{*}{AdvQA} & \fpthirtytwo{} & 18 & 0 & 0 & 0 & 0 & 0 & 0 & 0 & 0 \\
 &  & \fpsixteen{} & 19 & 0.13 & 0.27 & 0 & 0.03 & $-0.03$ & 0.03 & 0.07 & 0 \\
 &  & \bfsixteen{} & 19 & 0.77 & 1.00 & 0.17 & 0.23 & $-0.17$ & 0 & 0.13 & 0 \\
\midrule[0.5pt]
\multirow{9}{*}{DeBERTa-v3-L (NLI)} & \multirow{3}{*}{MNLI} & \fpthirtytwo{} & 18 & 0 & 0 & 0 & 0 & 0 & 0 & 0 & 0 \\
 &  & \fpsixteen{} & 19 & 0 & 0 & 0 & 0 & 0 & 0 & 0 & 0 \\
 &  & \bfsixteen{} & 19 & 0.05 & 0.10 & 0 & 0 & 0 & 0.05 & 0.10 & 0 \\
\arrayrulecolor{black!30}\cline{2-12}\arrayrulecolor{black}
 & \multirow{3}{*}{SNLI} & \fpthirtytwo{} & 18 & 0 & 0 & 0 & 0 & 0 & 0 & 0 & 0 \\
 &  & \fpsixteen{} & 19 & 0 & 0 & 0 & 0 & 0 & 0 & 0 & 0 \\
 &  & \bfsixteen{} & 19 & 0 & 0.05 & 0 & 0.05 & $-0.05$ & 0 & 0.05 & 0 \\
\arrayrulecolor{black!30}\cline{2-12}\arrayrulecolor{black}
 & \multirow{3}{*}{ANLI} & \fpthirtytwo{} & 18 & 0 & 0 & 0 & 0 & 0 & 0 & 0 & 0 \\
 &  & \fpsixteen{} & 19 & 0 & 0.06 & 0 & 0.01 & 0 & 0 & 0.05 & 0 \\
 &  & \bfsixteen{} & 19 & 0.15 & 0.15 & 0.05 & 0.15 & $-0.15$ & 0.05 & 0.15 & 0 \\
\midrule[0.5pt]
\multirow{4}{*}{Wav2Vec2-base} & \multirow{2}{*}{LibriSpeech clean} & \fpthirtytwo{} & 14 & 0 & 0.27 & 0 & 0 & 0 & 0 & 0 & 0 \\
 &  & \bfsixteen{} & 17 & 23.44 & 40.39 & 0.39 & 0.78 & $-0.39$ & $-0.39$ & 0 & 0 \\
\arrayrulecolor{black!30}\cline{2-12}\arrayrulecolor{black}
 & \multirow{2}{*}{LibriSpeech other} & \fpthirtytwo{} & 14 & 0 & 0.27 & 0 & 0 & 0 & 0 & 0 & 0 \\
 &  & \bfsixteen{} & 17 & 35.94 & 44.14 & 0 & 0.39 & 0 & 0.39 & 7.81 & 0 \\
\midrule[0.5pt]
\multirow{6}{*}{Whisper-small} & \multirow{3}{*}{LibriSpeech clean} & \fpthirtytwo{} & 14 & 0 & 0 & 0 & 0 & 0 & 0 & 0 & 0 \\
 &  & \fpsixteen{} & 17 & 0 & 0.39 & 0 & 0 & 0 & 0 & 0.39 & 0 \\
 &  & \bfsixteen{} & 17 & 1.17 & 3.52 & 0 & 0.78 & $-0.39$ & 0 & 0.39 & 0 \\
\arrayrulecolor{black!30}\cline{2-12}\arrayrulecolor{black}
 & \multirow{3}{*}{LibriSpeech other} & \fpthirtytwo{} & 14 & 0 & 0 & 0 & 0 & 0 & 0 & 0 & 0 \\
 &  & \fpsixteen{} & 17 & 0.39 & 1.25 & 0 & 0 & 0 & 0 & 0 & 0 \\
 &  & \bfsixteen{} & 17 & 5.08 & 6.25 & 0.78 & 1.56 & $-1.56$ & $-0.78$ & 0 & 6 \\
\bottomrule
\end{tabular*}
\end{table*}

\section{Additional Controlled Kernel Perturbation Results}
\label{app:perturbation-results}
\setcounter{figure}{0}

\begin{figure*}[!p]
\centering
\includegraphics[width=\textwidth]{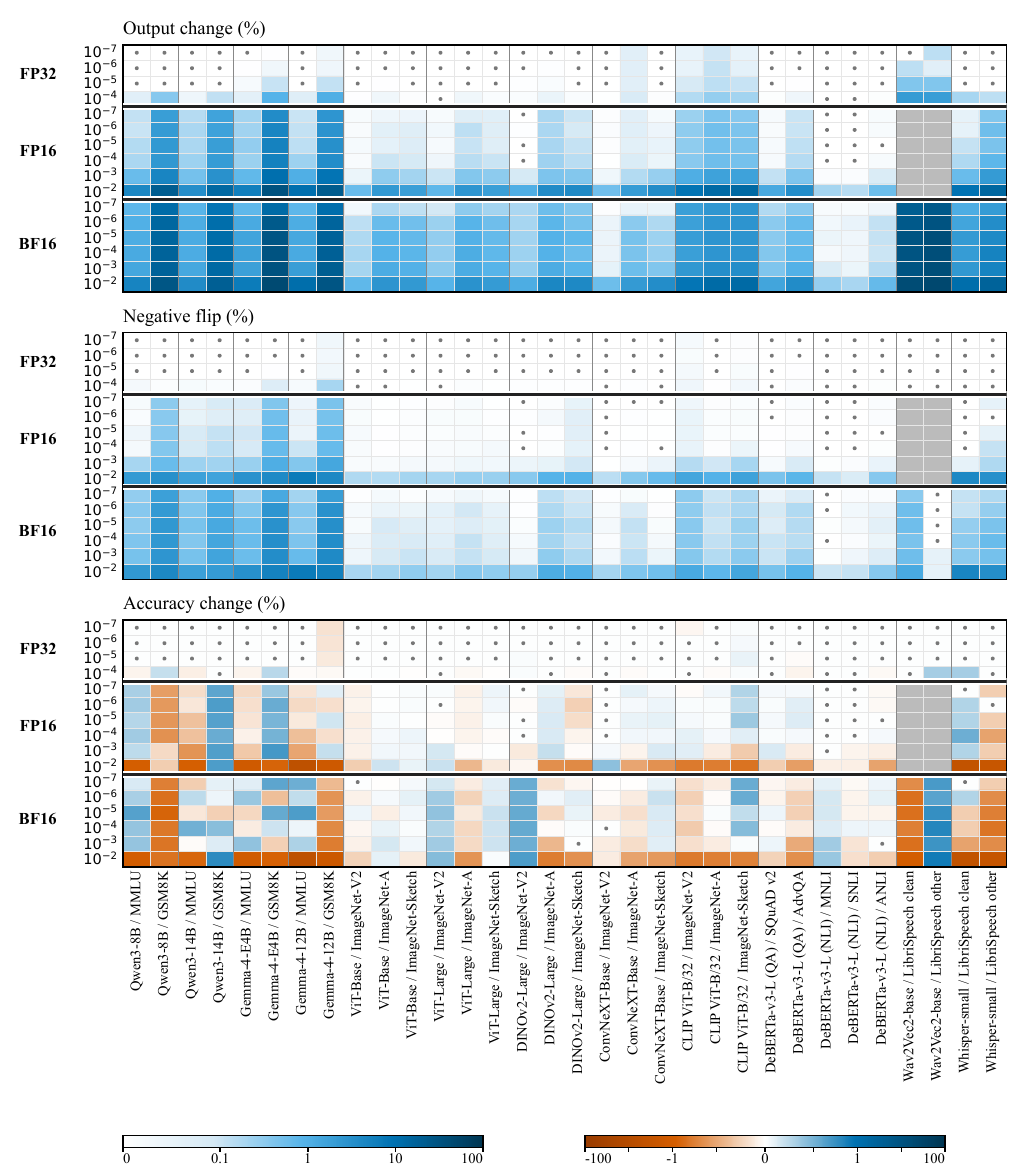}
\caption{Controlled kernel perturbation results across models and datasets. Panels show $R_{\mathrm{change}}$, $R_{\mathrm{flip}}$, and $\Delta_{\mathrm{acc}}$ in \%. Each panel contains \fpthirtytwo, \fpsixteen, and \bfsixteen blocks; rows show increasing perturbation upper bounds, $\varepsilon$. Each cell averages the evaluated operators. Dots mark exact zeros; gray cells mark combinations without evaluations. Blue indicates positive rates or accuracy gains; orange indicates accuracy drops.}
\label{app:perturbation-overview}
\end{figure*}

\figref{app:perturbation-overview} summarizes all 6,928 perturbation
evaluations. The three panels show $R_{\mathrm{change}}$,
$R_{\mathrm{flip}}$, and $\Delta_{\mathrm{acc}}$, respectively. Columns
identify model--dataset combinations. Each panel contains a block for each
numerical precision, with rows showing increasing perturbation upper bounds,
$\varepsilon$.

\para{Aggregation across operators.}
Each cell pools the event counts and model-input counts across evaluated
operators for the same model, numerical precision, dataset, and
$\varepsilon$. All operators in a cell use the same model inputs, so the
pooled rate equals the arithmetic mean of their rates. Accuracy change is
averaged in the same way. The pooled denominator counts model inputs once
per operator and therefore includes repeated inputs.
These averages can hide differences between operators, including accuracy
gains and drops that offset each other. Results for individual operators
are retained in the accompanying data.

\para{Perturbation settings.}
We use seed 0 and sample an independent sign and magnitude for each output
element, with magnitudes drawn uniformly from
$[\varepsilon/10,\varepsilon]$. The largest evaluated $\varepsilon$ is
$10^{-4}$ for \fpthirtytwo and $10^{-2}$ for \fpsixteen and \bfsixteen.
Comparisons across precisions therefore use the upper bounds evaluated
at all compared precisions.

\para{Figure scales.}
The two rate panels share a blue color scale. For accuracy change, orange
indicates a drop, white indicates zero, and blue indicates a gain.
The rate scale is linear from 0\% to 0.1\%; the accuracy-change scale is
linear within $\pm0.1\%$. Both scales are logarithmic outside these intervals.
Dots mark exact zeros, and gray cells mark combinations without evaluations.
All plotted values are unsmoothed averages.

\makeatletter
\newcommand{\appKipTableCaption}{\def\@captype{table}}
\makeatother
\twocolumn[
\begin{minipage}{\textwidth}
\section{KIP Training and Prediction Results}
\label{app:kip}
\setcounter{table}{0}
{\appKipTableCaption
\centering
\caption{KIP prediction performance across 21 budget pairs (all values in \%). Each budget row averages five balanced repetitions of 1,000 sampled replacement evaluations. Recall$_{>0.5}$ is the fraction of evaluations with accuracy drops greater than 0.5\% that KIP flags. The last columns give the fraction not flagged and, among those evaluations, the fraction exceeding either observed budget.}
\label{app:kip-results}
\fontsize{8.5}{10.5}\selectfont
\setlength{\tabcolsep}{3pt}
\begin{tabular*}{\textwidth}{@{\extracolsep{\fill}}cccccccc@{}}
\toprule
\multicolumn{2}{c}{Budget (\%)} & \multirow{2}{*}{F1} & \multirow{2}{*}{Precision} & \multirow{2}{*}{Recall} & \multirow{2}{*}{Recall$_{>0.5}$} & \multirow{2}{*}{Not flagged} & \multirow{2}{*}{\shortstack[c]{Budget violations\\among not flagged}} \\
\cmidrule(lr){1-2}
Output change & Negative flip & & & & & & \\
\midrule
\multirow{1}{*}{0.1} & 0.1 & 81.49 & 79.72 & 83.36 & 86.69 & 47.70 & 17.45 \\
\midrule[0.4pt]
\multirow{2}{*}{0.2} & 0.1 & 81.47 & 80.49 & 82.48 & 88.38 & 48.76 & 17.97 \\
 & 0.2 & 81.41 & 80.84 & 82.00 & 84.46 & 49.28 & 18.25 \\
\midrule[0.4pt]
\multirow{3}{*}{0.5} & 0.1 & 80.78 & 79.55 & 82.04 & 85.40 & 48.44 & 18.53 \\
 & 0.2 & 83.42 & 81.00 & 86.00 & 87.33 & 46.90 & 14.91 \\
 & 0.5 & 82.41 & 80.81 & 84.08 & 88.80 & 47.98 & 16.57 \\
\midrule[0.4pt]
\multirow{4}{*}{1} & 0.1 & 82.11 & 80.59 & 83.72 & 87.28 & 48.04 & 16.93 \\
 & 0.2 & 84.90 & 80.93 & 89.28 & 90.39 & 44.84 & 11.95 \\
 & 0.5 & 83.63 & 79.90 & 87.76 & 90.73 & 45.08 & 13.52 \\
 & 1 & 82.91 & 79.14 & 87.08 & 88.02 & 44.98 & 14.32 \\
\midrule[0.4pt]
\multirow{5}{*}{2} & 0.1 & 82.62 & 79.38 & 86.16 & 87.90 & 45.72 & 15.11 \\
 & 0.2 & 87.16 & 81.83 & 93.24 & 85.19 & 43.02 & 7.86 \\
 & 0.5 & 88.09 & 84.66 & 91.84 & 87.19 & 45.74 & 8.91 \\
 & 1 & 87.59 & 82.72 & 93.08 & 87.08 & 43.72 & 7.91 \\
 & 2 & 88.29 & 83.52 & 93.64 & 86.65 & 43.94 & 7.23 \\
\midrule[0.4pt]
\multirow{6}{*}{5} & 0.1 & 83.29 & 80.21 & 86.64 & 86.67 & 45.98 & 14.52 \\
 & 0.2 & 85.95 & 79.99 & 92.88 & 89.12 & 41.92 & 8.49 \\
 & 0.5 & 84.99 & 85.96 & 84.04 & 79.19 & 51.12 & 15.60 \\
 & 1 & 84.17 & 84.94 & 83.44 & 68.53 & 50.88 & 16.25 \\
 & 2 & 84.40 & 89.38 & 79.96 & 62.61 & 55.26 & 18.12 \\
 & 5 & 85.21 & 90.71 & 80.36 & 61.10 & 55.70 & 17.60 \\
\midrule
\multicolumn{2}{c}{Mean over 21 budget pairs} & 84.11 & 82.20 & 86.34 & 83.75 & 47.38 & 14.19 \\
\bottomrule
\end{tabular*}
}
\vspace{8pt}
\end{minipage}
]

\para{Training \& prediction.}
We train KIP on all 6,928 perturbation evaluations to jointly predict
$R_{\mathrm{change}}$ and $R_{\mathrm{flip}}$. Inputs are task type,
numerical precision, operator category, and the logarithm of the perturbation
upper bound; targets are the observed rates. The random forest has
120 trees, maximum depth 8, and at least 6 examples per leaf. Examples are
weighted by model-input count. We double the negative-flip target during
fitting and rescale it for prediction.
For real replacements, the measured maximum absolute kernel-level discrepancy
replaces the perturbation bound. Predictions use cumulative maxima over
61 logarithmically spaced magnitudes in the corresponding training range.
Between points, we use the next higher grid point; outside the range,
the nearest boundary. Zero discrepancy receives zero predicted rates.

\para{Evaluation protocol.}
The 2,739 replacement evaluations share the 1,299 inference configurations
of the perturbation study. KIP flags a replacement if either predicted rate
exceeds its budget; equality does not trigger a flag. The positive class
for precision, recall, and F1 comprises replacements exceeding either
observed budget.
Each of 21 budget pairs uses five balanced repetitions of 1,000 sampled
replacement evaluations: 500 that exceed either budget and 500 that exceed
neither. Sampling preserves accuracy-drop groups within each observed
budget class and weights inference configurations equally within each group.
Sampling master seed 9 was selected retrospectively from 128 candidates.
These repetitions resample existing evaluations; they do not estimate
deployment prevalence.

A perfect predictor leaves 50\% unflagged on these balanced samples;
false flags and missed budget violations can yield the same fraction.

\para{Results across budget pairs.}
F1 ranges from 80.78\% to 88.29\%. The $(2\%,2\%)$ pair has the highest F1
and the lowest fraction of budget violations among unflagged evaluations, 7.23\%.
At an output-change budget of 5\%, increasing the negative-flip budget
from 0.2\% to 5\% raises precision from 79.99\% to 90.71\%, while recall
falls from 92.88\% to 80.36\%. The fraction not flagged rises from
41.92\% to 55.70\%. The $(5\%,5\%)$ pair has the lowest
Recall$_{>0.5}$, 61.10\%; every missed evaluation with an accuracy drop
greater than 0.5\% still exceeds an observed budget.
Budget pairs differ in positive classes and sampled evaluations, so these
comparisons do not isolate the effect of changing a budget on a fixed test set.

\clearpage

\end{document}